\documentclass[accepted]{uai2026} 
                        
\usepackage[american]{babel}

\usepackage{natbib} 
\usepackage{mathtools} 
\usepackage{booktabs} 
\usepackage{tikz} 
\usepackage[utf8]{inputenc} 
\usepackage[T1]{fontenc}    
\usepackage{hyperref}       
\usepackage{url}            
\usepackage{booktabs}       
\usepackage{amsfonts}       
\usepackage{nicefrac}       
\usepackage{microtype}      
\usepackage{xcolor}         
\usepackage{algorithm}      
\usepackage{algorithmic}
\usepackage{graphicx}
\usepackage{amsmath}
\usepackage{tcolorbox}
\usepackage{multirow}
\usepackage{longtable}
\usepackage[table,xcdraw]{xcolor}
\usepackage{subcaption}

\newcommand{\swap}[3][-]{#3#1#2} 

\title{CIExplainer++: Generating Causal and Interpretable Explanations for Graph Neural Networks}

\author[1]{\href{mailto:<f.caldas@campus.fct.unl.pt>}{Francisco Caldas}{}}
\author[1]{Sahil Satish Kumar}
\author[1]{Ruben Belo}
\author[1]{Cláudia Soares}
\affil[1]{%
    NOVA LINCS\\
    NOVA School of Science and Technology\\
    Lisbon, Portugal
}
  
\begin{document}
\maketitle

\begin{abstract}
  Explainable Artificial Intelligence aims to make black-box models more trustworthy by presenting, in a human-understandable manner, the elements that lead to the model's output. This involves both (i) identifying components and connections with genuine causal influence on outputs and (ii) translating such structures into an interpretable representation. For the former, we introduce CIExplainer, a novel perturbation-based method grounded in causal inference for explaining Graph Neural Networks (GNNs). CIExplainer identifies the subgraph with the highest causal effects on GNN predictions using the Potential Outcome Framework. We evaluate and compare CIExplainer on various GNN architectures (GCN, GraphSAGE, GAT, GIN) and datasets.
  To bridge subgraph explanations with human interpretability, we further propose G2TeXplainer, a method that transforms causal subgraphs into natural language explanations that capture both feature-level and relational information.
\end{abstract}

\section{Introduction}
Graph Neural Networks (GNNs) have emerged as a powerful class of deep learning models for learning from graph-structured data~\cite{Zhang2024}. They have achieved strong performance across diverse domains, including recommendation systems \cite{gnnrecommendation1, gnnrecommendation2}, healthcare \cite{Duarte2021, Valdeira2023, gnnhealth1, kumar2024causal}, and crime prediction \cite{gnncrime1, gnncrime2}.

Despite their success, GNNs typically operate as black-box models (e.g., models that natively do not provide a clear understanding of the mechanisms that lead to a given prediction). This lack of transparency poses significant challenges in high-stakes environments, where understanding model predictions is essential for informed decision-making and increased trustworthiness and accountability. Consequently, developing principled methods to explain the GNN predictions has become a critical research direction. 

Explainable Artificial Intelligence (XAI)~\cite{Molnar2022,minh2022explainable} has emerged to address questions such as
\begin{quote}
    How can we 
\textbf{trust} a model prediction? How can we know \textbf{why} the model made the wrong prediction? How can we \textbf{explain} the model behavior?
\end{quote}

 XAI provides explanation methods to clarify the model's inner mechanisms. However, many explanation methods rely on identifying associations and correlations in the data to justify predictions. While these explanations are often transparent and easy to understand, they do not always offer a full picture of how a model works because ``correlation does not imply causation''. This means that associations and correlations may overlook deeper, cause-and-effect relationships that drive the models' predictions. 

To address this limitation, we propose \textbf{CIExplainer}, a local perturbation-based~\cite{Yuan2023, Zhang2024} causal explanation framework grounded in the Potential Outcome Framework. CIExplainer estimates the causal effect of feature-level interventions on model predictions, enabling the identification of graph components that have a causal impact on GNN outputs.
GNNs take graphs as input and produce embeddings for graph elements, which are then used for tasks such as node classification, graph classification, or link prediction. As is common in GNN explanation methods, the goal of CIExplainer is to produce an explanation subgraph. 
The explanation subgraph is nonetheless limited in interpretability, as it requires an expert to understand its structure and features~\cite{GraphXAIN}. 
\begin{figure*}[!ht]
    \centering
    \includegraphics[width=0.95\linewidth]{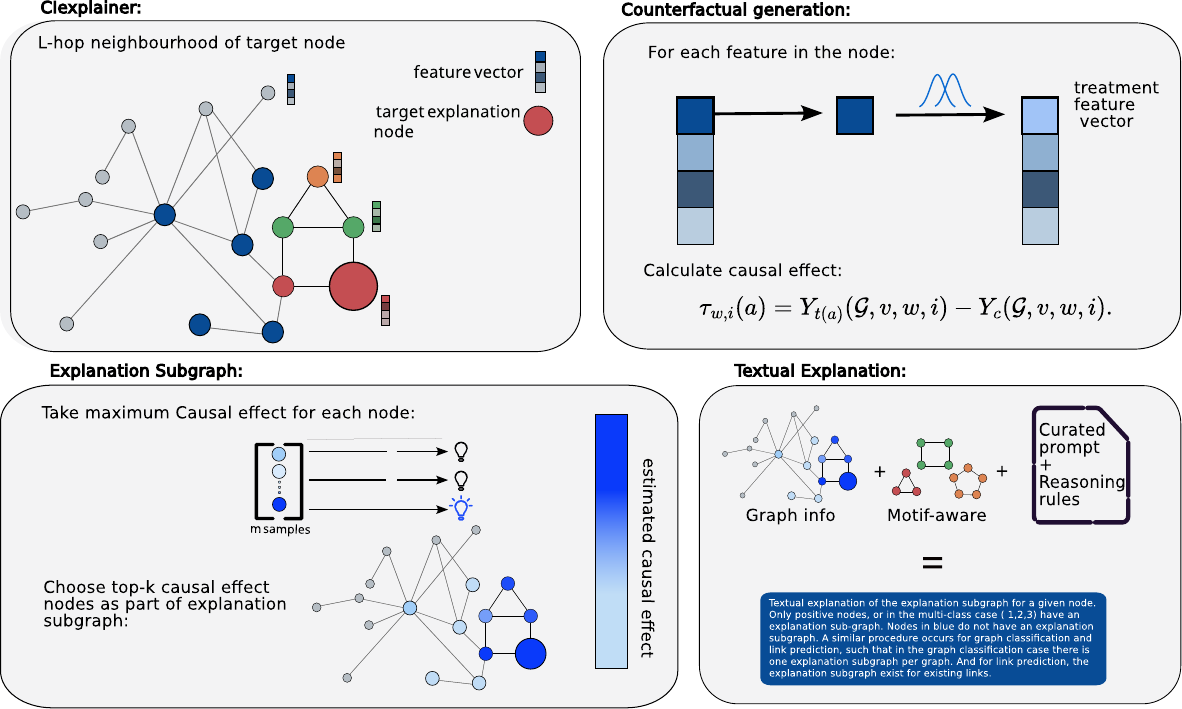}
    \caption{Diagram of the CIExplainer proposed pipeline. Using as example the explanation of a node classification task, CIExplainer++ outputs the nodes with highest Causal Effect, alongside a textual description of the explanation subgraph. }
    \label{fig:diagram}
\end{figure*}

To bridge this gap, we further propose \textbf{G2TeXplainer}, a graph-to-text model that translates causal subgraphs into natural language explanations. Given the explanation subgraph produced by CIExplainer and the key structural motifs within the graph we use a Large Language Model (LLM) to verbalize them in a way that is accessible to a broader audience without sacrificing technical rigor, directly addressing the questions posed above: now we can \textbf{understand} why a prediction was made, \textbf{trust} the reasoning behind it, and \textbf{explain} it. To evaluate the quality of the generated explanations, we design an automated LLM-as-judge~\cite{llm_as_judge} evaluation framework that assesses the best prompting strategy for generating faithful and clear explanations.

All code and datasets are available at \href{https://github.com/FranciscoCaldas/CIExplainer}{https://github.com/FranciscoCaldas/CIExplainer}

\section{Related Work}

\textbf{Graph Neural Networks.} The primary goal of GNNs is to perform representation learning on graphs, which means that they create a condensed representation of a graph, node, or edge in a lower-dimensional space while preserving the original relational information of the input graph. This condensed representation, or embedding, can then be applied in subsequent tasks, such as graph classification and link prediction. Graph Convolution Network (GCN) \cite{Kipf2017}, GraphSAGE \cite{Hamilton2017}, Graph Attention Network (GAT) \cite{Velic2018}, and Graph Isomorphism Network (GIN) \cite{Xu2019} are some of the most prominent GNN models to date. \textbf{GCN} \cite{Kipf2017} is a variant of convolutional neural networks that works directly on graphs, using a localized first-order approximation of spectral graph convolutions \cite{Hammond2011}. The model uses a graph convolutional operation that combines a node's features with those of its immediate neighbors, allowing it to learn representations sensitive to local graph structure. \textbf{GraphSAGE} \cite{Hamilton2017} is designed to generate node embeddings from large graphs. It accomplishes this by sampling a $K$-hop neighborhood around the target node and aggregating the features of the target node and its neighbors. Through this process, GraphSAGE learns a series of aggregator functions that, once trained, can generate embeddings for previously unseen nodes. \textbf{GAT} \cite{Velic2018} leverages masked self-attentional layers that allow nodes to focus on their neighbors' features with different weights. The \textbf{GIN} model \cite{Xu2019} is based on the Weisfeiler-Lehman test \cite{Leman1968}, which is well known for its ability to differentiate many types of graph structures by iteratively gathering information from neighboring nodes. The main idea of GIN is to use a very expressive aggregation function that can model injective functions. This allows the model to fully capture the set of node features in a graph neighborhood. 

The field of GNN models is extensive and a comprehensive survey is available in \citep{Wu2021_a}. Other relevant and recent models include GCN-II \cite{chen2020}, Anti-Symmetric Deep Graph Network (ADGN) \cite{gravina2023adgn}, Graph Neural Diffusion (GRAND) \cite{chamberlain2021grand}, Substructure Assembling Network - Soft Sequence with Context Attention (SAN-SSCA) \cite{yang2023} and Interpretable and Efficient Heterogeneous
Graph Convolutional Network (ie-HGCN) \cite{yang2021interpretable}.

\textbf{GNN Explanation.} Several perturbation-based methods have been proposed to provide instance-level explanations for GNN predictions. \textbf{GNNExplainer} \cite{Ying2019} is a local explanation method applicable to graph, node, and edge level tasks. Given a trained GNN and its prediction, it returns a compact explanation subgraph and a subset of node features that predominantly influence the output by maximizing the mutual information between the prediction and candidate subgraphs. However, it must be retrained for each instance. \textbf{PGExplainer} \cite{Luo2020,pge2024} similarly learns discrete edge masks by training a mask predictor to maximize mutual information between original and masked predictions, but is trained once and generalizes across examples. GRAPHMASK \cite{Schlichtkrull2021} is a post-hoc method that learns to identify removable edges while preserving predictive performance using stochastic gates optimized with $L_0$ regularization to encourage sparsity, yielding a graph that can be seen as an explanation graph. \textbf{SubgraphX} \cite{Yuan2021} combines Monte Carlo Tree Search and Shapley values to search for important connected subgraphs through iterative node pruning, where Shapley values quantify each subgraph's contribution to the prediction. Unlike other approaches, SubgraphX guarantees connected explanation subgraphs. 
GEM \cite{wanyu2021} proposes a generative explanation framework that learns to produce compact explanation subgraphs guided by an objective inspired by Granger causality. The method quantifies the importance of graph components by measuring changes in prediction error when edges or nodes are removed, and trains a graph generator to distill subgraphs that preserve model outputs while enforcing structural constraints such as connectivity and sparsity.

Other methods for local post-hoc, agnostic, GNN explanation mostly include surrogate models, such as PGM-Explainer \cite{Vu2020}, GraphLIME~\cite{Huang2023}, DnX \cite{pereira23a}, RC-Explainer \cite{Wang2022ReinforcedCE}, GOAt \cite{lu2024goat} and GstarX~\cite{zhang2022gstarx},that learn a different, white-box model, to explain the predictions of the original model. Also, and while not specifically designed for GNNs, gradient based methods \cite{shrikumar2017learning,Shrikumar2016NotJA} and saliency maps \cite{simonyan2013deep} can determine the importance of the input by evaluating the gradient of the model output with respect to the input features.

Furthermore, there exist methods that explicitly learn counterfactual explanation graphs~\cite{Lucic2022}, meaning producing the minimal subgraph with modified edges such that a prediction is meaningfully changed, or that focus on regression based tasks \cite{zhang2024regexplainer}. 

\textbf{Graph-to-Text Explanation.} While most GNN explanation methods focus on identifying important subgraphs or feature attributions, translating these structural explanations into natural language remains relatively underexplored. GraphXAIN~\cite{GraphXAIN} generates narrative explanations by prompting an LLM with explanatory subgraphs and feature importance scores produced by GNNExplainer, improving perceived understandability and user satisfaction. For text-attributed graphs, GraphNarrator~\cite{GraphNarrator} adopts a verbalization-first strategy: it converts saliency-based node and edge importance scores into structured textual descriptions and uses them to train a dedicated explanation generator. In contrast, LOGIC~\cite{Logic} aligns internal GNN embeddings with the LLM token space via a learned projector, constructing hybrid prompts that combine graph structure, node text, and latent representations. 
However, both GraphNarrator and LOGIC are tailored to text-attributed graphs, limiting their applicability to settings where nodes lack textual features. 

\section{Methodology}

One of the goals of the work presented in this paper is to construct an explanation method tailored to produce causal explanations for GNN-based tasks. As such, we formulate CIExplainer, a local explanation method dedicated to generating causal inference explanations. With CIExplainer, we provide a causal subgraph, that can, paired with its textual interpretation, improve the quality and transparency of GNN-based models, enabling their trustworthy deployment in critical decision-making environments like clinical risk modeling and financial fraud detection. In the following, we will develop the methodology in the context of node prediction, but it is trivially extended to link prediction and graph classification.

\subsection{Problem Statement}
\label{subsec:problem_statement}
Let $\mathcal{G} = (\mathcal{V}, \mathcal{E})$ denote a graph with a node set $\mathcal{V}$ and an edge set $\mathcal{E}$. Each node $v \in \mathcal{V}$ is associated with a feature vector $\mathbf{x} = \{x_1, x_2, ...x_n\}$, $x_i \in \mathbb{R}$. Let $\Phi$ denote a trained GNN model for a particular dataset $\mathcal{D}$. 

Inspired by \cite{Ying2019}, we emphasize that the computation graph of a node $v$ used for node prediction is defined by the neighborhood-based aggregation of the GNN. This graph fully determines all the information the GNN uses to compute a node-level prediction $\hat{y}$ for $v$. Specifically, the computation graph of $v$ guides the GNN in creating the embedding for $v$. Let $N\!e_K(v) = (\mathcal{V}_{Ne}, \mathcal{E}_{Ne}), \mathcal{V}_{Ne} \subseteq \mathcal{V}, \ \mathcal{E}_{Ne} \subseteq \mathcal{E}$ represent the computation graph of $v$, which is a $K$-hop sampled neighborhood around $v$. 

The prediction of a GNN is given by $\hat{y} = \Phi(N\!e_K(v))$, meaning it depends entirely on the model $\Phi$, the structural information of the graph $N\!e_K(v)$, and the related node features. This observation shows that to explain $\hat{y}$, we only need to focus on the sampled neighborhood $N\!e_K(v)$ and the corresponding node features. Thus, CIExplainer returns as an explanation a subgraph $\mathcal{G}_{EXP} \subset N\!e_K(v)$ with the nodes that had the highest causal effect on $\hat{y}$. The causal effect of each node on $\hat{y}$ is calculated using the Potential Outcome Framework \cite{Holland1986}.

\subsection{Potential Outcome Framework}\label{subsec:potential_outcome_framework}

The Neyman-Rubin Causal Model \cite{Holland1986}, also known as the Potential Outcome Framework, formalizes causal inference in terms of unit-level counterfactual outcomes. Let $U$ denote a population of units, where each unit $u \in U$ is the basic object of study. Variables are real-valued functions defined over $U$, assigning measurements to each unit.

The central concept in the model revolves around the potential to either expose or refrain from exposing each unit
to a causal influence. Let $S(u) \in {t, c}$ denote a binary treatment assignment, where $t$ represents treatment and $c$ control. For each unit $u$, define the potential outcomes $Y_t(u)$ and $Y_c(u)$ as the responses that would be observed under treatment $t$ and control $c$, respectively. The unit-level causal effect is defined as:

\begin{equation}
    \label{eq:causal_effect}
    CE = Y_t(u) - Y_c(u).
\end{equation}

According to the framework, \eqref{eq:causal_effect} expresses that treatment $t$ causes the effect $Y_t(u) - Y_c(u)$ on unit $u$ (relative to treatment $c$).  

\subsection{CIExplainer}\label{subsec:ciexplainer}

\paragraph*{Causal Estimand.} For a fixed trained GNN~$\Phi$, a target prediction instance~$(\mathcal{G},v)$, a candidate node~$w$ in the graph and a feature~$i$ define the unit as the explanation instance associated with~$(\mathcal{G},v,w,i)$. The control condition~$c$ keeps the original feature value~$x_{w,i}$. The treatment condition~$t(a)$ replaces~$x_{w,i}$ by an admissible intervention value~$a$, while keeping the trained model fixed. The outcome is the prediction score, logit, or probability assigned by~$\Phi$ to the target class. Then,
\begin{equation}
\label{eq:tau}
\tau_{w,i}(a) = Y_{t(a)}(\mathcal{G},v,w,i) - Y_{c}(\mathcal{G},v,w,i). 
\end{equation}
For continuous features,  several counterfactual values are sampled from $q_i(a)$, the chosen intervention distribution for feature~$i$. Define the intervention values as an iid batch of size $B$ $a_{w,i}^{(1)}, \dots, a_{w,i}^{(b)}, \dots, a_{w,i}^{(B)} \overset{\text{iid}}{\sim} q_i$. Define the sample-level aggregation operator $$\mathop{\text{Agg}}_{b=1,\dots,B} \: : \: \mathbb{R}^{B} \to \mathbb{R},$$
that can be the average, the sum, or the point maximum. Then, the finite-budget estimand is
\begin{equation}
\tau_{w,i}^{\mathrm{Agg}, B}
=
\mathbb{E}_{\left\{a_{w,i}^{(b)}\right\}_{b=1}^B \overset{\text{iid}}{\sim} q_i}
\left[
\mathop{\text{Agg}}_{b=1,\dots,B} 
\tau_{w,i} \left (a_{w,i}^{(b)}\right )
\right],
\end{equation}
where $\tau_{w,i}(a)$ was defined in~\eqref{eq:tau}.

The parameter $B$ is a parameter of the algorithm and depends on the computational budget available. The current implementation uses $B=10$ and $\mathop{\text{Agg}} = \max$, yielding a finite-budget maximum-effect score. Other choices, such as mean, maximum absolute effect, or median, correspond to different intervention-effect summaries and are assessed in ablations.
In practice, with one sampled batch, the empirical score for max-aggregation is
$$
\hat{\tau}_{w,i}^{B} = \max_{b=1,\cdots,B} \left\{ Y_{t\left (a_{w,i}^{(b)}\right )}(\mathcal{G},v,w,i) - Y_{c}(\mathcal{G},v,w,i) \right\}.
$$

\paragraph*{Explanation Subgraph} Given a node $v$ extracted from a graph \( \mathcal{G} \), a trained GNN model \( \Phi \), and a node prediction probability \( \hat{y}_p \) generated by \( \Phi \) for that particular node by sampling a \( K \)-hop neighborhood \( N\!e_K = (\mathcal{V}_N, \mathcal{E}_N) \) of the node, CIExplainer yields a subgraph \( \mathcal{G}_{EXP} = (\mathcal{V}_{EXP}, \mathcal{E}_{EXP}) \) as an explanation for \( \hat{y} \), denoting the inferred node label by analyzing \( \hat{y}_p \). This subgraph contains the top \( l \) nodes and the edges between those nodes that caused \( \Phi \) to output \( \hat{y}_p \). Specifically, \( \mathcal{V}_{EXP} \subseteq \mathcal{V}_N \), \( \mathcal{E}_{EXP} \subseteq \mathcal{E}_N \), and \( |\mathcal{V}_{EXP}| = l \), where \( l \in \{1, 2, ..., | \mathcal{V}_N|\} \) is a hyperparameter enforcing the returned explanation to be concise and informative. Our proposed method, CIExplainer, selects the top \( l \) nodes by evaluating the causal effect that each node in \( N\!e_K \) exerts on the prediction \( \hat{y} \), then prioritizing the \( l \) nodes with the highest causal effect values. This computation of causal effects leverages the potential outcome framework for causal inference at the unit level, which in our case are the individual features of the node. 

Let \( \hat{y} \) be the node classification of the GNN for node $v$ and
\( S \) be a binary variable representing the cause, \( t \) or \( c \), to which \( \hat{y} \) is exposed. We note that to generate a node classification probability \( \hat{y}_p \), a GNN model \( \Phi \) only needs the sampled \( K \)-hop neighborhood \( N\!e_K \) from node \( v \). As such, manipulating \( \hat{y} \) consists of manipulating \( N\!e_K \). Hence, let \( c \) denote the absence of manipulation of \( N\!e_K \) and let \( t \) denote the manipulation of \( N\!e_K \), specifically, manipulating the feature values of a node in \( N\!e_K \). Maintaining the original \( N\!e_K \) used to generate \( \hat{y} \) is denoted by \( S(u) = c \) and it represents the \textit{actual} outcome, while, manipulating the features of nodes in \( N\!e_K \) is denoted by \( S(u) = t \) and it represents the \textit{counterfactual} outcome. Let \( Y \) denote the node prediction probability generated by \( \Phi \) for some sampled \( K \)-hop neighborhood. Then, \( Y_c(u) \) denotes the node prediction probability when \( u \) is exposed to \( c \), that is, it denotes the original or \textit{actual} node prediction probability \( \hat{y}_p \). Whereas \( Y_t(u) \) denotes the node prediction probability when \( u \) is exposed to \( t \), that is, the \textit{counterfactual} node prediction probability obtained by constructing a \textit{counterfactual} \( K \)-hop neighborhood.

In this context, the \textbf{Temporal Stability} holds because the output produced by the model \( \Phi \) remains consistent regardless of when the input is provided to \( \Phi \). Essentially, the value \( Y_c(u) \) remains unchanged irrespective of the timing of exposing \( c \) to \( u \) and measuring \( Y_c(u) \). Similarly, the \textbf{Causal Transience} assumption is valid because exposing \( c \) to \( u \) does not alter the overall network structure of \( N\!e_K \), and computing the output of \( \Phi \) for \( N\!e_K \) does not change the weights of the \( \Phi \). Consequently, when computing the output for the original network configuration \( N\!e_K \), denoted as \( Y_t(u) \), prior exposure of \( u \) to \( c \) does not influence the result. Therefore, the causal effect \( CE \) of altering the feature value of a node in \( N\!e_K \) on the predicted outcome \( \hat{y} \), as assessed by \( Y \), can be determined using \eqref{eq:causal_effect}.

Given the aforementioned input conditions, and considering each node possesses \( n \) features, our method CIExplainer generates an explanation in accordance with Alg.~\ref{alg:CIExplainer}.


\begin{algorithm}
    \caption{CIExplainer for node prediction}
    \label{alg:CIExplainer}
    \begin{algorithmic}
        \STATE {\bfseries Input:} target node \( v \), GNN model \( \Phi \), predicted label \( \hat{y}_v \)
        \STATE {\bfseries Output:} explanation subgraph \( \mathcal{G}_{EXP} \) containing the nodes that caused \( \hat{y}_v \)
        \STATE
        \STATE Sample a \( K \)-hop neighborhood \( N\!e_K(v) \) around node \( v \)
        \STATE Compute the node prediction \( \hat{y}_v \) using \( \Phi \) on \( N\!e_K(v) \)
        \FOR{each node \( w \in N\!e_K(v) \)}
                \FOR{each feature \( x_i \) of \( w \)}
                    \STATE Generate a counterfactual by perturbing \( x_i \)
                    \STATE Compute the causal effect \( CE_i \) of the perturbation on \( \hat{y}_v \)
                \ENDFOR
            \STATE Node causal effect \(CE_u = \max_i |CE_i| \)
        \ENDFOR
        \STATE Return subgraph of \( N\!e_K(v) \) containing top \( l \) nodes by \( CE_u \) and edges between those nodes
    \end{algorithmic}
\end{algorithm}

For discrete (binary or categorical) features, counterfactual nodes are generated via direct value intervention, replacing the observed feature with an alternative admissible category (for binary variables, its complement).


For continuous features, let \( \mathbf{x}_w = (x_{w,1}, \dots, x_{w,d}) \) denote the feature vector of node \( w \).
To generate a counterfactual with respect to feature \( x_{w,i} \), we hold all remaining features fixed and replace
\( x_{w,i} \) with a value sampled from its empirical marginal distribution,
$x_{w,i}' \sim p(x_i),
$
yielding the intervened feature vector
\[
\mathbf{x'}_{w} = (x_{u,1}, \dots x_{u,i}', \dots, x_{u,d}).
\]
The distribution \( p(x_i) \) is estimated from the training data. In our experiments, we impose a fixed per-feature sampling budget of 10 counterfactual perturbations per feature.

This algorithm is extended for the tasks of node classification and graph classification by modifying the K-hop Neighborhood that defines the set of nodes that have a causal effect on the outcome.

\textbf{Link Prediction.} For the link prediction task of an edge between a pair of nodes \( v, w \), we adjust CIExplainer by sampling the \( K \)-hop neighborhood $N\!e_K(v)$ around \( v,w \) while keeping the rest of the algorithm unchanged.

\textbf{Graph Classification.} For the graph classification task of a graph $\mathcal{G}$, we modify CIExplainer to use the entire graph $\mathcal{G}$ instead of a sampled neighborhood $N\!e_K$.

\subsection{G2TeXplainer}

Given an explanation subgraph $\mathcal{G}_{EXP}$ and the neighborhood $N\!e_K$ the goal is to provide a textual explanation for a node classification task, and more generally for any task. Following the work in GraphXAIN~\cite{GraphXAIN} we also include a textual explanation of the GNN task. 
The prompt can be further enriched with structural properties of the subgraph and the original graph, such as node degree, $k$-cycle structures, cliques, and graph density. The effectiveness of the LLM depends on how systematically this structural information is incorporated. Providing irrelevant details may degrade explanation quality, whereas insufficient structural context can increase hallucinations or lead to vague, uninformative responses.
Overall, whereas prior work relies on structured human user studies assessing subjective dimensions (e.g., understandability, trust, and satisfaction), we use an LLM-based judge to assess explanation correctness and faithfulness to the input and extracted subgraph, enabling a more task-oriented and reproducible evaluation.

\section{Experimental Studies}

First, we describe the datasets, the baseline explanation methods we compared, and the details of the experimental settings. Then, we present the results of explaining different GNN model architectures on various datasets and tasks.

\begin{table}[t]
    \caption{Dataset statistics. The characteristics are defined per graph. BA-Shapes and Tree-Grid contain a single graph, with multiple motifs attached, where each motif is a ground-truth explanation.}
    \label{tab:datasets}
    \centering
    \small
    \setlength{\tabcolsep}{4pt}
    \begin{tabular}{c c c c c}
        \toprule
         & BA-Shapes & T-Grids & BA-2motifs & MUTAG \\ 
        \midrule
        
        
        Motif 
        & \includegraphics[height=1cm]{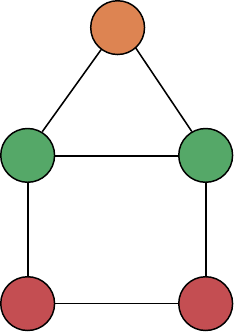}
        & \includegraphics[height=1cm]{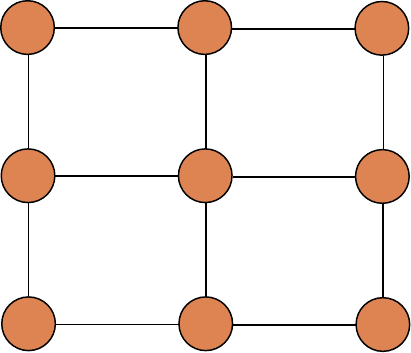}
        & \includegraphics[height=1cm]{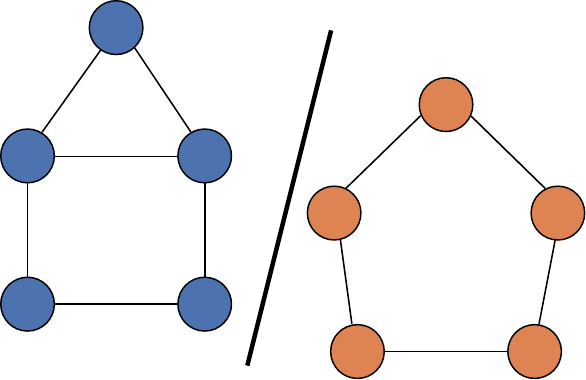}
        & \includegraphics[height=1cm]{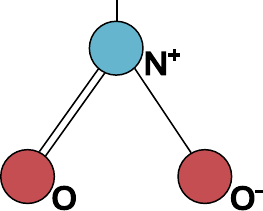} \\
        
        \midrule
        
        Graphs & 1 & 1 & 1000 & 4337 \\
        Nodes & 700 & 1231 & 25 & 418 \\
        Edges & 2055 & 1565 & $\approx$ 26 & $\approx$ 225 \\
        Features & 2 & 2 & 2 & 14 \\
        Classes & 4 & 2 & 2 & 2 \\
        Avg. Degree & 5.87 & 2.54 & 2.04 & 1.07 \\
        Density & 0.0084 & 0.0021 & 0.0849 & 0.0026 \\ 
        \bottomrule
    \end{tabular}
\end{table}

\textbf{Datasets.} We follow and extend PGExplainer \cite{Luo2020} experimental settings and use two synthetic explanation evaluation datasets for node classification: BA-Shapes and Tree-Grids; one synthetic explanation evaluation dataset and one real-world dataset for graph classification: BA-2motif and MUTAG; 
Table \ref{tab:datasets} shows the details of synthetic and real-world datasets. The \textbf{BA-Shapes} dataset, described in \cite{Ying2019}, is a single graph created using a base Barabasi-Albert (BA) graph with 300 nodes. To this base graph, 80 ``house''-shaped motifs are attached to nodes selected randomly. After attaching these motifs, additional random edges are added to introduce some perturbation in the graph. In BA-Shapes, nodes do not have features. Nodes from the base BA graph are labeled as 0, while nodes located at the top, middle, and bottom of the "house" motifs are labeled as 1, 2, and 3, respectively. In the \textbf{Tree-Grids} dataset, the base graph is an 8-level balanced binary tree. To this base graph, 80 motifs of 3x3 grids are attached randomly. The \textbf{BA-2motifs} dataset from \cite{Luo2020}, includes 800 graphs. Each graph starts with a BA base graph. Half of the graphs are attached with ``house'' motifs, while the other half are attached with five-node cycle motifs. The type of motif attached determines the class label of the graph, making it a binary classification task. In these datasets, the ground truth explanations are artificially designed to correspond to the motifs that nodes belong to. 

The \textbf{MUTAG} dataset consists of 4,337 molecule graphs, each classified into one of two categories based on its mutagenic effect. Each node has a one-hot encoded feature vector of size 14, indicating the chemical element of the node. It is known~\cite{Luo2020} that carbon rings with chemical groups like $NH_2$ or $NO_2$ have a mutagenic effect. As noted in \cite{Luo2020}, carbon rings are present in both mutagenic and non-mutagenic graphs, so they are not discriminative. Therefore, we consider carbon rings as shared base graphs, while $NH_2$ and $NO_2$ are treated as motifs for the mutagenic graphs, serving as ground-truth explanations. Non-mutagenic graphs, however, do not have clear motifs. In this setup, we train different GNN models on the whole dataset but evaluate explanations only for predictions on mutagenic graphs since they have ground-truth explanations. We define a positive explanation as one that identifies one mutagenic group completely, regardless of the total number of mutagenic groups in that specific graph. 

Considering how CIExplainer works, we extend the synthetic datasets that originally lack node features by assigning features to the nodes. To make GNN models use these features for predictions, we introduce node features artificially by using different Gaussian distributions with varying means for each node class while keeping the variance fixed at $2$. For a dataset with \(C\) classes, each node’s feature vector is sampled as \(x \sim \mathcal{N}(\mu_c, 2)\), where \(c \in \{0,1,...,C\}\). 

\textbf{Baselines.} For comparison we choose more widely known GNN-specific explanaiers, and  explanation methods that can be generally applied to any prediction model. Thus, we compared our method with GNNExplainer \cite{Ying2019}, PGExplainer \cite{Luo2020} (which is a surrogate model that generates perturbations), and SubgraphX \cite{Yuan2021}. And also, with gradient based methods IntegratedGradient \cite{shrikumar2017learning}, InputXGradient \cite{Shrikumar2016NotJA} and Shapley Values \cite{Shapley1951Notes}, we also compare with a random explainer \cite{Fey2019}, which is a simple explainer that makes random explanations, thus it can be used as a basic reference for all other methods.

\textbf{Metrics.} For every prediction \(\hat{y} = \Phi(\mathcal{G}_c)\) from the test set of the dataset \(\mathcal{D}\) made by a GNN \(\Phi\) on a computation graph \(\mathcal{G}_c\) , we calculate an explanation subgraph \(\mathcal{G}_{EXP}\). Then, we use all the calculated explanation subgraphs on the test set to evaluate the explanation method using various metrics. We use the \textbf{Jaccard Index} to measure the overlap between the ground-truth explanation nodes and the \(\mathcal{G}_{EXP}\) nodes. Formally, given the set of nodes $\mathcal{V}_T$ from a ground-truth explanation graph and the set of nodes $\mathcal{V}_{EXP}$ from an explanation subgraph, the Jaccard index, also known as the intersection over union ($IoU$), is defined as:
\begin{equation*}
  IoU(\mathcal{V}_T, \mathcal{V}_{EXP}) = \frac{\vert \mathcal{V}_T \cap \mathcal{V}_{EXP} \vert}{\vert \mathcal{V}_T \cup \mathcal{V}_{EXP} \vert}
\end{equation*}
A good explanation method should have a high Jaccard index, indicating that the explanation subgraph is similar to the ground-truth explanation. For a direct comparison with other works~\cite{Ying2019}, we also evaluate the models on \textbf{Precision}. 
Given a ground-truth explanation $\mathcal{G}_T$ and $\mathcal{G}_{EXP}$, Precision ($Pr$) is defined as:
\begin{equation*}
    Pr(\mathcal{G}_T, \mathcal{G}_{EXP}) = \frac{TP(\mathcal{G}_T, \mathcal{G}_{EXP})}{TP(\mathcal{G}_T, \mathcal{G}_{EXP}) + FP(\mathcal{G}_T, \mathcal{G}_{EXP})}\\ 
\end{equation*}
and 

where $TP(\mathcal{G}_T, \mathcal{G}_{EXP})$ is the number of true positives, $FP(\mathcal{G}_T, \mathcal{G}_{EXP})$ is the number of false positives. 
We obtain the \textbf{Inference Time} of each test example and report its average across the test set on the result tables. For a fair comparison, we explictly present the training time for PGExplainer separately. 

\begin{table}[t]
\centering
\caption{Computational cost comparison. Train time is reported once per explainer, with inference cost average per model and dataset.
Inference time is averaged per explained node. All experiments run on a 10GB partition of a single NVIDIA A100 GPU.}
\label{tab:inference_time}
\footnotesize
\setlength{\tabcolsep}{4pt}
\renewcommand{\arraystretch}{1.05}
\begin{tabular}{lcc}
\toprule
Method 
& Train Time (sec.) 
& Inference (sec./node)  \\
\midrule
Random Exp. & -- & 0.0003 $\pm$ 0.0  \\
PGExplainer      & 88.2 & 0.0025 $\pm$ 0.00005 \\
\midrule
IntegratedGradient & -- & 0.258 $\pm$ 0.0235 \\
InputXGradient & -- & 0.214  $\pm$ 0.0236\\
\midrule
Shapley Values & -- & 38.064 $\pm $0.790 \\
GNNExplainer     & -- & 1.314  $\pm$ 0.0304 \\
SubgraphX        & -- & 7.866 $\pm$ 0.512  \\
CIExplainer      & -- & \textbf{0.553} $\pm$ 0.0113\\
\bottomrule
\end{tabular}
\end{table}

\begin{table*}[t]
\centering
\caption{Explanation results for node classification. Standard deviation estimated from 10 independent runs, with sampling and training repeated for each run. Best result in \textbf{bold}, second best is \underline{underlined}. }
\label{tab:explainers_stacked}
\footnotesize
\setlength{\tabcolsep}{3pt}
\renewcommand{\arraystretch}{1.05}
\begin{tabular}{llcccccccc}
\toprule

& Models 
& \multicolumn{2}{c}{GCN} 
& \multicolumn{2}{c}{GAT} 
& \multicolumn{2}{c}{GIN} 
& \multicolumn{2}{c}{GraphSAGE} \\
\cmidrule(lr){3-4} \cmidrule(lr){5-6} \cmidrule(lr){7-8} \cmidrule(lr){9-10}
& & \bfseries IoU ($\uparrow$) & \bfseries Pr ($\uparrow$)
& \bfseries IoU ($\uparrow$) & \bfseries Pr ($\uparrow$)
& \bfseries IoU ($\uparrow$) & \bfseries Pr ($\uparrow$)
& \bfseries IoU ($\uparrow$) & \bfseries Pr ($\uparrow$) \\
\midrule
\multirow{5}{*}{\rotatebox[origin=c]{90}{\textit{BA-shapes}}}
& Random Explainer
& 0.0029$\pm$.01&0.0052$\pm$.01 
& 0.0037$\pm$.00&0.0062$\pm$.01 
& 0.0032$\pm$.00&0.0057$\pm$.01 
& 0.0040$\pm$.00&0.0071$\pm$.00 \\
\cmidrule(lr){2-10}
& IntegratedGradient
&  0.6590$\pm$.00& 0.7714$\pm$.00
& 0.7902$\pm$.00&0.8619$\pm$.00
& 0.5961$\pm$.00&0.7238$\pm$.00 
& \underline{0.8424$\pm$.00}&\underline{0.9000$\pm$.00} \\
& InputXGradient
& 0.7021$\pm$.00& 0.8000$\pm$.00
& \underline{0.8299$\pm$.00}&\underline{0.8857$\pm$.00}
& 0.5848$\pm$.00&0.7143$\pm$.00 
& 0.8424$\pm$.00&\underline{0.9000$\pm$.00} \\
& Shapley Values
& 0.6107$\pm$.03&  0.7343$\pm$.02
& 0.7249$\pm$.03& 0.8195$\pm$.02
& 0.5228$\pm$.03&0.6633$\pm$.03
& 0.7370$\pm$.02&0.8314$\pm$.01 \\
\cmidrule(lr){2-10}
& PGExplainer
& \underline{0.8163$\pm$.05}&\underline{0.8710$\pm$.10}
&0.5731$\pm$.05&0.6395$\pm$.06 
& \textbf{0.8566$\pm$.03}&\textbf{0.8929$\pm$.02} 
& 0.6378$\pm$.14&0.7210$\pm$.12 \\
& GNNExplainer 
&   0.5683$\pm$.02&0.7019$\pm$.01 
& 0.2922$\pm$.01&0.3967$\pm$.01 
& 0.5372$\pm$.03&0.6614$\pm$.02
& 0.4487$\pm$.03&0.5976$\pm$.02 \\
& SubgraphX
& 0.5363$\pm$.01&0.6729$\pm$.01
& 0.5935$\pm$.03& 0.7010$\pm$.02    
& 0.7369$\pm$.02&0.8181$\pm$.02
& 0.6036$\pm$.02&0.7338$\pm$.01 \\
& \textbf{CIExplainer (Ours)} 
    & \textbf{0.8909$\pm$.02} & \textbf{0.9314$\pm$.01 }
& \textbf{0.8448$\pm$.01} &\textbf{ 0.8738$\pm$.01} 
& \underline{0.7915$\pm$.02} & \underline{0.8667 $\pm$.01 }
& \textbf{0.9294$\pm$.01} &\textbf{ 0.9576$\pm$.04} \\
\midrule
\multirow{5}{*}{\rotatebox[origin=c]{90}{\textit{Tree-grid}}}
& Random Explainer 
& 0.0036$\pm$.00 & 0.0067$\pm$.00
& 0.0053$\pm$.00 & 0.0099$\pm$.01
& 0.0040$\pm$.00 & 0.0075$\pm$.00
& 0.0024$\pm$.00 & 0.0046$\pm$.00 \\
\cmidrule(lr){2-10}
& IntegratedGradient
&  0.8868$\pm$.00& 0.9361$\pm$.00
& \underline{0.8395$\pm$.00}&\underline{0.8976$\pm$.00}
&\textbf{ 0.8741$\pm$.00}&\textbf{0.9285$\pm$.00 }
& 0.8868$\pm$.00&0.9361  $\pm$.00 \\
& InputXGradient
& \underline{0.8887$\pm$.00}   &\underline{0.9376$\pm$.00}
&  0.8381$\pm$.00& 0.8954$\pm$.00
& 0.0182$\pm$.00&0.0228$\pm$.00
& \underline{0.8890$\pm$.00}&\underline{0.9376  $\pm$.00} \\
& Shapley Values
& 0.7375$\pm$.01&  0.8356$\pm$.01
& 0.7453$\pm$.02& 0.8444$\pm$.01
& 0.1072$\pm$.00&0.1813$\pm$.01
& 0.7487$\pm$.02&0.8440$\pm$.01 \\
\cmidrule(lr){2-10}
& PGExplainer 
& 0.0971$\pm$.17 & 0.1335$\pm$.20 
& 0.0333$\pm$.00 & 0.0580$\pm$.02 
& 0.8701$\pm$.02 & 0.9253$\pm$.01 
& 0.1960$\pm$.06 & 0.1960$\pm$.07 \\
& GNNExplainer 
& 0.7754$\pm$.01&0.8658$\pm$.01
& 0.7413$\pm$.01&0.8382$\pm$.00
& 0.0182$\pm$.00&0.0228$\pm$.00 
& 0.7449$\pm$.01&0.8397$\pm$.00 \\
& SubgraphX
& 0.2985$\pm$.01&0.4464$\pm$.01  
& 0.3042$\pm$.01&0.4534$\pm$.02 
&  0.3118$\pm$.01&0.4632$\pm$.02 
& 0.3124$\pm$.02&0.4647$\pm$.02  \\
& \textbf{CIExplainer (Ours)}
& \textbf{0.9008$\pm$.01}&\textbf{0.9381$\pm$.01}  
& \textbf{0.8439$\pm$.00}&\textbf{0.9075$\pm$.01}  
&  \underline{0.8407$\pm$.00}&\underline{0.8988$\pm$.00} 
& \textbf{0.9325$\pm$.02}&\textbf{0.9595$\pm$.01}  \\
\bottomrule
\end{tabular}
\end{table*}


\textbf{Experiment Details.} 
We run each explanation method with each GNN model on each dataset 10 times and report the average results with the standard deviation. We experiment with four types of GNN models: GCN \cite{Kipf2017}, GraphSAGE \cite{Hamilton2017}, GAT \cite{Velic2018}, and GIN \cite{Xu2019}. We use the default hyperparameters of all explanation methods. 

\subsection{G2TeXplainer Experimental Setup}

\textbf{Prompt Design.}
We propose four prompt variants that progressively increase structural richness and reasoning guidance. 
\textbf{P1 (Graph)} provides the raw graph edges without explicit reasoning instructions.
\textbf{P2 (1-hop)} retains only the 1-hop neighborhood of the explanation subgraph, including edges among selected nodes and their immediate neighbors.
\textbf{P3 (1-hop + R.)} retains P2’s input structure while adding explicit reasoning rules to focus on selected nodes and the predicted motif.
Finally, \textbf{P4 (Motif-Aware + R.)} replaces raw edge lists with detected motifs (e.g., triangles, 4- and 5-cycles) while retaining P3’s structured reasoning, aligning the representation with motif-level reasoning.

\textbf{Generation Configurations.} To generate natural language explanations, we use LLaMA 3 8B~\cite{llama3} with stochastic decoding via nucleus sampling. We set the temperature to 0.7 to balance coherence and variability, and use a top‑p of 0.9 to restrict token selection to high-probability candidates. A mild repetition penalty of 1.1 is applied to improve fluency and reduce redundancy. Random seeds are controlled to ensure reproducibility across samples while allowing some variation between generations. For each example, we produce 30 explanations across 20 \textbf{BA-2motifs} instances. In addition, we evaluate on the full test set using a reduced sampling budget of 5 explanations per instance; the corresponding results are reported in Appendix~\ref{app:g2t_extra_eval}.


\textbf{Evaluation Criteria.}
We evaluate each natural language explanation using five complementary criteria:
\begin{enumerate*}[label=(\arabic*)]
\item \textbf{Node Fidelity}, which evaluates whether the explanation assigns structural importance only to the ground-truth subgraph nodes, and avoids overclaiming importance to other nodes. 
\item \textbf{Structure}, which evaluates whether the explanation accurately describes the detected subgraph motifs (e.g., cycles, triangles) and correctly identifies which nodes belong to each motif, without inventing structures;
\item \textbf{Clarity}, which evaluates how coherently and unambiguously the explanation communicates the reasoning behind why the selected nodes support the GNN's prediction;
\item \textbf{Semantic Similarity}, which assesses consistency across independent runs and is computed as the average of TF-IDF lexical similarity and embedding-based semantic similarity; and
\item \textbf{Pairwise Preference} which also relies on an LLM-as-judge protocol, where two explanations for the same instance are compared with randomized order to avoid positional bias and the judge selects the preferred explanation~\cite{llm_uncertain_uai}.
\end{enumerate*}
The LLM-as-judge evaluations are conducted using Gemma 3 27B~\cite{gemma3}, which scores the relevant criteria on a scale from 1 (very poor) to 5 (perfect)~\cite{scoresLabels}.

In addition to the automatic evaluations, we conduct a human evaluation study. Explanations are assessed using four binary factual criteria (\textbf{Correct Class}, \textbf{Correct Nodes}, \textbf{Correct Structure}, and \textbf{No Hallucination}), which are summed into a \textbf{Factual Consistency Score}, as well as two subjective criteria, \textbf{Clarity} and \textbf{Overall Quality}, rated on a 1--5 Likert scale.

\subsection{Quantitative Results}

The results for node classification are reported in Table \ref{tab:explainers_stacked}. Across both datasets, CIExplainer consistently identifies the correct explanation subgraph across all backbone models, with stability across multiple runs. In contrast, PGExplainer exhibits high variability depending on the underlying GNN, with the largest standard deviations, indicating unstable precision across runs. Although never outperforming other methods, SubgraphX remains stable across datasets and models, never collapsing for any (dataset, model) pair.

While the considered tasks are relatively simple, GIN performs poorly on the Tree-Grid dataset (see Table \ref{tab:nc_test_metrics} in Supplement \ref{sec:training_appendix}), which significantly degrades the explanation accuracy of GNNExplainer. This behavior aligns with the intuition that explanation quality is bounded by the predictive performance of the underlying model.

Overall, CIExplainer produces more accurate explanation subgraphs with consistent behavior across architectures. Although its performance reflects the predictive quality of the backbone GNN, it does not collapse when the predictor is imperfect. On the BA-shapes dataset, approximately 40\% of CIExplainer explanations achieve a perfect match (IoU = 1) with the ground-truth motif. In terms of inference time, CIExplainer is significantly faster than comparable methods that do not rely on offline training, as can be seen in Table \ref{tab:inference_time}. We exclude the inference cost of G2TeXplainer from this comparison, as other explainers do not include a graph-to-text interpretation component.

The graph classification results are reported in Table \ref{tab:graph_explain} in the Supplementary Material \ref{sec:gc_exp}. On BA-2motif, CIExplainer achieves the best or second-best performance across most backbone architectures, remaining competitive for all models with the exception of GAT.
On MUTAG, results are more heterogeneous. GNNExplainer achieves the strongest performance for GCN, GAT, and GraphSAGE, while CIExplainer performs best under the GIN backbone. We note that GIN is the model with the best performance on the MUTAG dataset, as seen in Table~\ref{tab:gc_test_metrics}, and that we found a positive correlation between the explanation and the model quality, for the CIExplainer. 

Overall, CIExplainer demonstrates  strong performance on in node classification tasks, and comparable performance with other explainers in graph classification, at a reduced computational cost.

\begin{figure*}[!h]

    \centering
    
    \begin{subfigure}{0.48\textwidth}
        \centering
        \includegraphics[width=\linewidth]{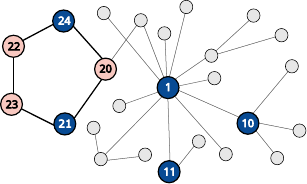}
        \label{fig:ba_shapes}
    \end{subfigure}
    \hfill
    \begin{subfigure}{0.48\textwidth}
        \centering
        
        \includegraphics[width=\linewidth]{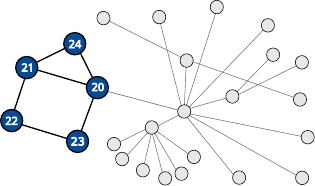}
        \label{fig:tgrid}
    \end{subfigure}
    
    \vspace{0.5em}
    
    \begin{subfigure}{0.48\textwidth}
        \centering
        \includegraphics[width=\linewidth]{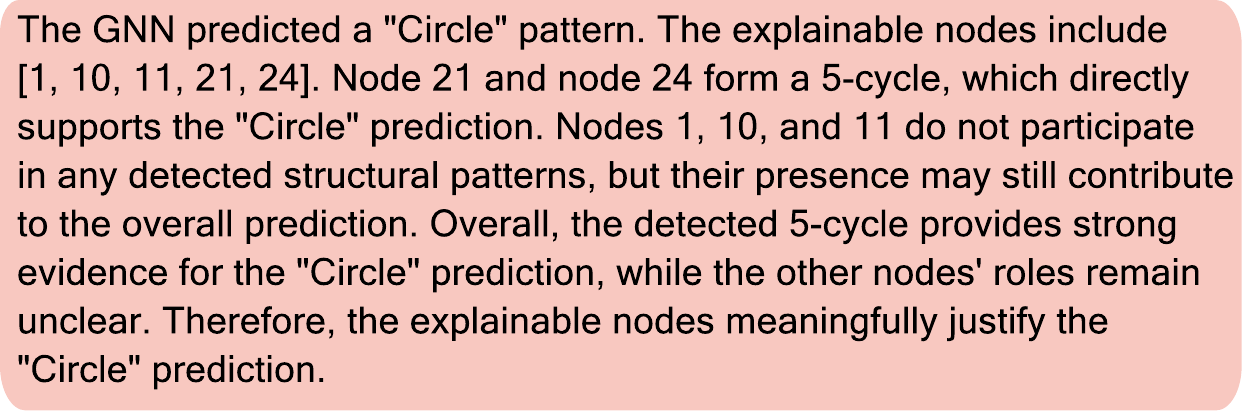}
        \caption{}
        \label{fig:ba_2motifs}
    \end{subfigure}
    \hfill
    \begin{subfigure}{0.48\textwidth}
        \centering
        \includegraphics[width=\linewidth]{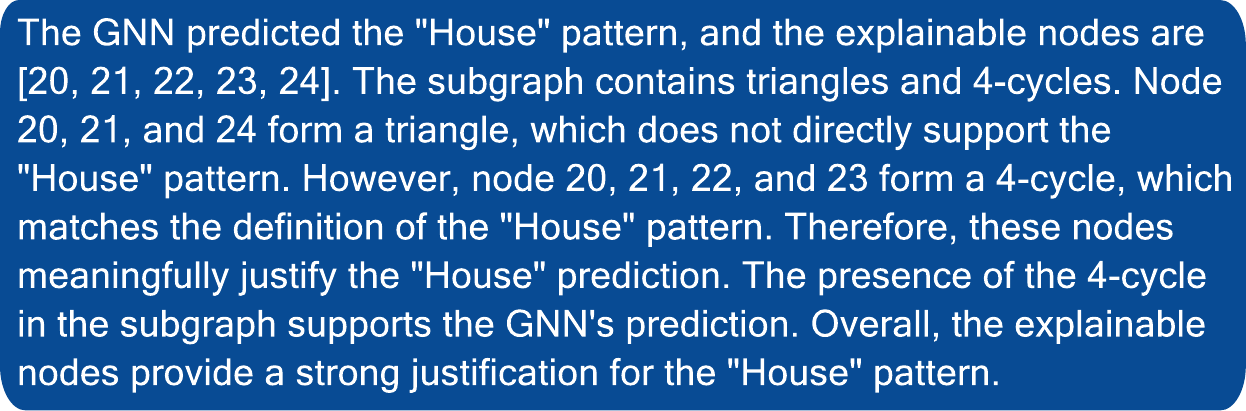}
        \caption{}
        \label{fig:mutag}
    \end{subfigure}
    
    \caption{Two examples of the explained subgraph alongside the textual description, for the BA-2Motifs dataset.}
    \label{fig:dataset_examples}
\end{figure*}
\subsection{Qualitative Results}

\begin{table*}[th]
\centering
\caption{Evaluation of G2TeXplainer and GraphXAIN on a subset of the test set (20 examples, 30 generated explanations per example). Judge-based metrics are scored from 1 (very poor) to 5 (perfect). \textbf{N. Fid.}: correctness of node importance attribution; \textbf{Str.}: accuracy of the identified motifs and node memberships; \textbf{Clar.}: coherence and interpretability of the explanation. Human evaluation metrics include \textbf{FacC.} (Factual Consistency Score, computed from Correct Class, Correct Nodes, Correct Structure, and No Hallucination), \textbf{Clar.} (human-rated clarity), and \textbf{Ovr.} (overall quality), both rated on a 1--5 Likert scale. Bold indicates the best result and \underline{underlined} values indicate the second-best result.
}
\label{tab:g2texplainer_llm_judge}
\footnotesize
\setlength{\tabcolsep}{2.7pt}
\renewcommand{\arraystretch}{1.05}

\begin{tabular}{llcccccccc}
\toprule

& \multirow{2}{*}{\textbf{Prompt}}
&
& \multicolumn{4}{c}{LLM-as-Judge}
& \multicolumn{3}{c}{Human Eval}
\\
\cmidrule(lr){4-7}
\cmidrule(lr){8-10}

&
& \textbf{Sim.}
& \textbf{Str.}
& \textbf{Clar.}
& \textbf{N. Fid.}
& \textbf{PW.\%}
& \textbf{FacC.}
& \textbf{Clar.}
& \textbf{Ovr.}
\\
\midrule
\multirow{4}{*}{\rotatebox[origin=c]{90}{\fontsize{8pt}{9.5pt}\selectfont GraphSAGE}}
& GraphXAIN Adap.     &  --     & 3.754 & 3.862 & 3.881 &        --          & \underline{4.095} & 3.143 & 3.190 \\
& P1 Graph            & 0.712 & 3.576 & 3.284 & 3.687 & \underline{40.2} & 4.000 & 3.381 & 2.810 \\
& P2 1-hop            & \underline{0.727} & 3.448 & 3.254 & \underline{3.697} & 38.8 & \textbf{4.333} & \underline{4.048} & \underline{3.476} \\
& P3 1-hop + R.       & 0.689 & \underline{3.770} & \underline{2.994} & 3.459 & 32.0 & 3.095 & 3.190 & 2.286 \\
& P4 Motif-Aware + R. & \textbf{0.751} & \textbf{4.143} & \textbf{4.340} & \textbf{3.763} & \textbf{89.0} & \textbf{4.333} & \textbf{4.238} & \textbf{3.857}\\
\midrule
\addlinespace[2.5pt]
\raisebox{-1.2ex}{\rotatebox{90}{\fontsize{8pt}{9.5pt}\selectfont GCN}} & P4 Motif-Aware + R. &  \textbf{0.758} & 4.262 & \underline{4.348} & \textbf{3.789} & -- \\
\addlinespace[2.5pt]
\midrule
\addlinespace[2.5pt]
\raisebox{-1.2ex}{\rotatebox{90}{\fontsize{8pt}{9.5pt}\selectfont GAT}} & P4 Motif-Aware + R. & 0.744 & \textbf{4.354} & \textbf{4.403} & 3.538  & -- \\
\addlinespace[2.5pt]
\midrule
\addlinespace[2.5pt]
\raisebox{-1.2ex}{\rotatebox{90}{\fontsize{8pt}{9.5pt}\selectfont GIN}} & P4 Motif-Aware + R. & \underline{0.754} & \underline{4.276} & 4.271 & 3.724 & -- \\
\addlinespace[2.5pt]
\bottomrule
\end{tabular}
\end{table*}

\textbf{Per-Metric Analysis.}
Table~\ref{tab:g2texplainer_llm_judge} reveals a clear progression across prompt variants within GraphSAGE. \textbf{Node Fidelity} and \textbf{Structure} improve substantially from P1 to P4, confirming that motif-aware prompting with explicit scope constraints reduces overclaiming and improves motif identification. \textbf{Clarity} and \textbf{Similarity} remain comparatively stable across all prompts and backbones, ranging narrowly between 3.459--3.789 and 0.689--0.758, indicating that fluency and cross-run consistency are less sensitive to prompt design than structural correctness. Across GNN backbones, P4 differences are small, suggesting that prompt design is the dominant driver of explanation quality.

Table~\ref{tab:g2texplainer_llm_judge} also reports \textbf{Pairwise Win-rates} obtained using an LLM-as-judge. Each percentage indicates how often the row prompt version was preferred over alternative prompts across matched explanation pairs. The results reveal a decisive advantage for the motif-aware prompt (P4). It is overall preferred over the other three prompts 89.0\%. In contrast, earlier prompt variants (P1--P3) exhibit near-balanced competition, with win rates bellow 50\%, indicating only modest differences.
The substantial improvement observed for P4 can be attributed to its explicit reasoning constraints. Unlike earlier prompts (P1--P3), which primarily refine wording and contextual information, P4 enforces strict scope limitations (e.g., restricting discussion to selected nodes). This design likely explains P4’s consistent preference across 6300 pairwise comparisons.

Human evaluation further corroborates these findings. P4 achieves the highest overall scores in both \textbf{Clarity} (4.238) and \textbf{Overall Quality} (3.857), while also attaining the joint-best \textbf{Factual Consistency Score} (4.333). These results suggest that the improvements identified by the LLM-as-judge translate to explanations that are not only preferred automatically, but are also perceived by human annotators as clearer, more accurate, and more useful.

In Fig. \ref{fig:dataset_examples} we have two curated examples for the BA-2Motif dataset using our proposed pipeline. A recurrent artifact is that CIExplainer tends to overvalue highly connected nodes, like node 1 in example (a). This is because, while the central node is not part of the motif, is still very relevant to the GNN, and a counterfactual feature in that node produces a significant impact on the model prediction. In the textual description, the text correctly identifies the nodes that are part of the 5-cycle, and is able to summarize the explanation subgraph. In example (b) we see a case where the model correctly identifies the complete "house" motif. Across the entire dataset, we found that house motifs had clearer and more faithful descriptions, as shown in Fig. \ref{fig:gcn_struct_score_distribution} in the Supplement, this can be due to the triangle pattern, since a ``house'' motif is defined by having a triangle, while a ``circle'' is defined by not having a triangle, which is more complex to describe. 

\section{Discussion and Limitations}
The algorithm used to calculate the feature-based causal effect has nonetheless some limitations, that we would like to point out: While generating a counterfactual to a binary feature is straightforward, for finite sets it requires defining a set of treatments $S = \{t_1,t_2,\dots,t_k\}$ where for $t_k \neq c$ we have multiple counterfactual outcomes $Y_{t_k}(u)$. Due to computational limitations, we employ a sampling scheme where the counterfactual treatment is chosen from the marginal probability of the feature, with Laplace smoothing. Similarly, for continuous variables, we leverage the feature marginal probability and enforce a sampling scheme that is biased towards treatments that are statistically significantly different from the control value. The use of the marginal distribution instead of the joint feature distribution is also due to the simplicity of the approach, and to guarantee scalability for high-dimensional data, where it is infeasible to estimate and to sample from such a joint distribution. 
After computing the Causal effect $CE_i$, on logits, over a set of features $\{x_i\}_{i=1}^k$, the CIExplainer takes the maximum causal effect as the node causal effect. This is a deliberate choice, since the goal is to identify the feature with the highest direct counterfactual dependence \cite{beckers2022}, not the node. We also like to acknowledge that our work is based on the Neyman-Rubin Causal Model and not the Pearl Causal Model \cite{pearl2009}.

\section{Conclusions}
We presented a framework for explanations of GNN models that transforms black-box predictions into explanation subgraphs paired with textual descriptions.
We conducted a robust evaluation of CIExplainer across multiple GNN architectures, datasets, and tasks, using diverse explanation metrics and comparisons against state-of-the-art GNN explainers. Specifically, for node classification, CIExplainer achieved the strongest overall performance while also exhibiting the lowest inference time among perturbation-based methods.
The key takeaway is that the Potential Outcome Framework can be used for causal inference and explanation, for both graph and node classification. Since there still a gap between real and synthetic datasets for GNN explanation, one research direction includes the generation of more complex, realistic datasets, that still have ground truth explanations, but with rich and meaningful feature spaces, that more realistically represented most use-cases for GNNs. This would enable us to better understand the strengths and weaknesses of different explanation methods and support the design of improved alternatives. 
Generated counterfactuals should not only achieve the desired effect but also be realistic and computationally efficient. As future work, we aim to improve the realism of the generated causal counterfactuals to better reflect plausible interventions.  
And, as GNN explanations gain adoption in real-world scenarios, calibration analysis becomes an important research direction. Furthermore, accurate textual descriptions, alongside feature-based explanations, will become more necessary to bridge the gap between ML experts and end users.

\bibliography{uai2026-template}

\onecolumn

\title{Supplementary Material}
\maketitle
\appendix

\section{Graph Classification Explaination Results}
\label{sec:gc_exp}

Table~\ref{tab:graph_explain} reports detailed explanation performance for graph classification across backbone architectures and datasets.

\textbf{BA-2motif.}
On this synthetic motif-based dataset, CIExplainer achieves the highest or second-highest IoU and precision for GCN, GIN, and GraphSAGE backbones. In particular, for GCN and GIN, CIExplainer attains the best overall performance. For GraphSAGE, CIExplainer remains competitive with PGExplainer. Across all GNN models, one key feature that can be seen in the examples shown in Fig. \ref{fig:dataset_examples} is that CIexplainer tends to identify nodes with high degree as relevant for the explanation.

\textbf{MUTAG.}
On the real-world MUTAG dataset, results vary depending on the underlying GNN. GNNExplainer achieves the strongest performance for GCN, GAT, and GraphSAGE, whereas CIExplainer performs best under the GIN backbone. The variability across backbones highlights that explanation quality remains dependent on the inductive biases of the underlying GNN model, particularly in more challenging datasets. 

Overall, these results suggest that causal subgraph identification is particularly effective when predictions are driven by localized structural motifs, as in BA-2motif, while remaining competitive on real-world molecular graphs. The observed backbone sensitivity further indicates that the interaction between aggregation mechanisms and causal perturbations warrants deeper investigation.

\begin{table*}[!h]
\centering
\caption{Explanation results for graph classification. Standard deviation obtained from 10 independents explanations.}
\label{tab:graph_explain}
\footnotesize
\setlength{\tabcolsep}{3pt}
\renewcommand{\arraystretch}{1.05}
\begin{tabular}{llcccccccc}
\toprule
& Models 
& \multicolumn{2}{c}{GCN} 
& \multicolumn{2}{c}{GAT} 
& \multicolumn{2}{c}{GIN} 
& \multicolumn{2}{c}{GraphSAGE} \\
\cmidrule(lr){3-4} \cmidrule(lr){5-6} \cmidrule(lr){7-8} \cmidrule(lr){9-10}
& & \bfseries IoU ($\uparrow$) & \bfseries Pr ($\uparrow$)
& \bfseries IoU ($\uparrow$) & \bfseries Pr ($\uparrow$)
& \bfseries IoU ($\uparrow$) & \bfseries Pr ($\uparrow$)
& \bfseries IoU ($\uparrow$) & \bfseries Pr ($\uparrow$) \\
\midrule

\multirow{5}{*}{\rotatebox[origin=c]{90}{\textit{BA-2motif}}}
&Random Explainer
& 0.1224$\pm$.01&0.2020$\pm$.01 
& 0.1234$\pm$.01&0.2036$\pm$.02 
& 0.1235$\pm$.01&0.2048$\pm$.02 
& 0.1171$\pm$.01&0.1942$\pm$.02 \\
\cmidrule(lr){2-10}
& IntegratedGradient
&  0.0022$\pm$.00& 0.0040$\pm$.00
& 0.0847$\pm$.00&0.1520$\pm$.00
& 0.0000$\pm$.00&0.0000$\pm$.00 
& 0.1652$\pm$.00&0.2700$\pm$.00 \\
& InputXGradients
&   0.0111$\pm$.00& 0.0200$\pm$.00
& 0.1225$\pm$.00&0.2120$\pm$.00
& 0.0078$\pm$.00&0.0140$\pm$.00 
& 0.2815$\pm$.00&0.4280$\pm$.00 \\
& Shapley Values 
& 0.0158$\pm$.01 & 0.0278$\pm$.01 
& 0.0115$\pm$.00 & 0.0206$\pm$.01
& 0.0161$\pm$.00 & 0.0288$\pm$.01 
& 0.0224$\pm$.00  &  0.0392$\pm$.01 \\
\cmidrule(lr){2-10}
&PGExplainer
& \underline{0.6369$\pm$.08}&\underline{0.7216$\pm$.05}
& \textbf{0.7696$\pm$.08}&\textbf{0.8272$\pm$.07}
& \underline{0.3859$\pm$.02}&\underline{0.5210$\pm$.02}
& \textbf{0.8823$\pm$.08}&\textbf{0.9240$\pm$.06} \\
&GNNExplainer 
&  0.1144$\pm$.01&0.1896$\pm$.02
& 0.0738$\pm$.01&0.1264$\pm$.02  
& 0.1056$\pm$.01&0.1772$\pm$.02
& 0.1129$\pm$.01&0.1880$\pm$.01 \\
&SubgraphX
& 0.3015$\pm$.02&0.3554$\pm$.03
& \underline{0.3108$\pm$.04}&\underline{0.3676$\pm$.03}
& 0.3301$\pm$.04&0.3864$\pm$.04
&0.3192$\pm$.03&0.3726$\pm$.03 \\
&\textbf{CIExplainer (Ours)} 
& \textbf{0.6745$\pm$.02}&\textbf{0.7842$\pm$.01}
&  0.1349$\pm$.00&0.2292$\pm$.01
& \textbf{0.3921$\pm$.01}& \textbf{0.5498$\pm$.01}
& \underline{0.7503$\pm$.01}&\underline{0.8336$\pm$.01} \\
\midrule

\multirow{5}{*}{\rotatebox[origin=c]{90}{\textit{MUTAG}}}
&Random Explainer 
&  0.1107$\pm$.01&0.1747$\pm$.02
& 0.1084$\pm$.00&0.1719$\pm$.01
& 0.1105$\pm$.01&0.1751$\pm$.02
& 0.1090$\pm$.01&0.1723$\pm$.02 \\
\cmidrule(lr){2-10}
& IntegratedGradient
&  \underline{0.4459$\pm$.00}& \underline{0.5580$\pm$.00}
& 0.1202$\pm$.00&0.1985$\pm$.00
& 0.1283$\pm$.00& 0.2040$\pm$.00
& \textbf{0.3025$\pm$.00}&\textbf{0.4246$\pm$.00} \\
& InputXGradients
&    0.0827$\pm$.00& 0.1100$\pm$.00
& 0.1208$\pm$.00&0.1748$\pm$.00
&  0.1281$\pm$.00&0.1982 $\pm$.00 
& 0.1097$\pm$.00&0.1683$\pm$.00 \\
& Shapley Values 
& 0.1884$\pm$.01 &0.2952$\pm$.01
& \textbf{0.1899$\pm$.01} & \textbf{0.2922$\pm$.01}
& \textbf{0.2138$\pm$.02}&  \textbf{0.3177$\pm$.02}
&\underline{0.2651$\pm$.02}  & \underline{ 0.3846$\pm$.02} \\
\cmidrule(lr){2-10}
&PGExplainer 
& 0.0780$\pm$.04&0.1200$\pm$.06 
& 0.1132$\pm$.03&0.1658$\pm$.00 
& 0.1000$\pm$.01&0.1549$\pm$.05
& 0.1650$\pm$.09&0.2343$\pm$.13 \\
&GNNExplainer 
& \textbf{0.6251$\pm$.02}&\textbf{0.7225$\pm$.01} 
& \underline{0.1659$\pm$.01}&\underline{0.2467$\pm$.02} 
&0.0611$\pm$.01&0.0950$\pm$.01 
&0.1872$\pm$.01&0.2871$\pm$.02 \\
&SubgraphX
& 0.0808$\pm$.01&0.1159$\pm$.01  
& 0.0877$\pm$.01&0.1259$\pm$.01 
&  0.0861$\pm$.01&0.1235$\pm$.01 
& 0.0863$\pm$.01&0.1240$\pm$.02  \\
&\textbf{CIExplainer (Ours)}
& 0.1160$\pm$.01&0.1844$\pm$.02
& 0.1314$\pm$.01&0.2141$\pm$.02
&   \underline{0.1751$\pm$.01}&\underline{0.2580$\pm$.01}
& 0.1213$\pm$.01 & 0.1890$\pm$.01   \\
\bottomrule
\end{tabular}
\end{table*}

\section{Counterfactual Generation}

As seen in \ref{alg:CIExplainer}, for each feature $x_i$, a counterfactual is generated, and the causal effect $CE_i$ is then computed. In this section we expand on the counterfactual generation process, which is non trivial, apart from binary features.

\subsection{Binary Features}

For binary features, the space of possible interventions is inherently discrete and limited to two states. Let $x_i \in \{0,1\}$ denote a binary a feature. The complement is $x'_i = 1-x_i$. Consequently, unlike continuous or multi-valued categorical features, binary features do not require sampling over a range of plausible values, as only a single alternative state exists.

However, in realistic settings, dependencies with other features and feasibility constraints may require additional adjustments, that distinguish intervenable feature level nodes from features where the intervention is not possible.

\subsection{Discrete  Features}

In this setting, the space of possible interventions is a finite set of discrete values. Let $ x_i \in \mathcal{X}_i$, where $ \mathcal{X}_i = \{v_1, \dots, v_K\} $ denotes the finite domain of the feature $i$. A feature-level counterfactual intervention corresponds to replacing the observed value $ x_i $ with an alternative value $ x_i' \in \mathcal{X}_i \setminus \{x_i\} $. 

While the space can exhaustively evaluated to estimate the maximum causal effect, doing so might be computationally infeasible for large sets, and higher number of features. Due to this limitation we employ a sampling scheme, that samples from the prior marginal distribution of the feature. Since, for the graph classification case, the prior information might be small and unrealiable, we perform laplace smoothing, mixing the the prior distribution with a uniform distribution. This guarantees that even in small datasets, all values can be sampled from. 

Following the same approach as in the binary feature case, not all features can receive an intervention. Moreover, changing just one feature might obtain an out-of-distribution feature vector. For small feature spaces, it is possible to sample from the empirical joint distribution, guaranteeing that the sample is typically found in the distribution, however the joint support size scales exponentially with the number of features, making it computationally costly to sample from this distribution. 

\begin{algorithm}[H]
\caption{Discrete Counterfactual with Laplace Smoothing}
\label{alg:dis}
\begin{algorithmic}[1]
\STATE \textbf{Input:} Feature index $j$, feature vector $\mathbf{x}$, marginal distribution $\mathcal{M}$, Domain $\mathcal{X}$,smoothing parameter $\epsilon$
\STATE \textbf{Output:} feature vector $\mathbf{x'}$ with conterfactual perturbation
\STATE obtain current value $v \leftarrow x_j$ 
\STATE valid values $\mathcal{V}' \leftarrow \mathcal{V}_j \setminus \{v\}$ 
\STATE new distribution $\mathbf{p}' \leftarrow \mathbf{p}$ without $v$
\FOR{each $k$ in $\mathbf{p}'$}
    \STATE $p'_k \leftarrow p'_k + \epsilon \frac{1}{|\mathcal{V}'|}$
\ENDFOR
\STATE Normalize $\mathbf{p}'$
\STATE Sample $v' \sim \text{Categorical}(\mathcal{V}', \mathbf{p}')$
\STATE $x_j \leftarrow v'$ 
\RETURN $\mathbf{x}$

\end{algorithmic}
\end{algorithm}

In Algorithm \ref{alg:dis} we showcase how the sampling is perform for one feature in the feature vector, using the marginal distribution. This algorithm is then used for each feature, and repeated $m$ times, from which the Causal Effect is extracted. 

\subsection{Continuous Features}
\label{sec:cont_feat}
In the continuous setting, the space of possible interventions is an uncountable set. Let $ x_j \in \mathcal{X}_j \subseteq \mathbb{R} $, where $ \mathcal{X}_j $ denotes the domain of the feature. A feature-level counterfactual intervention corresponds to replacing the observed value $ x_j $ with an alternative value \( x_j' \in \mathcal{X}_j \).

Unlike the discrete case, the intervention space cannot be exhaustively evaluated. As a result, estimating the maximum causal effect requires sampling-based approximations. In this work, we adopt a sampling scheme, where candidate counterfactuals are drawn from a continuous marginal distribution associated with the feature. For CIExplainer, this marginal distribution can be a bounded uniform distribution, an empirical Gaussian Distribution, a t-Student distribution with heavier tails, or Gaussian Mixture model.

To ensure sufficient and efficient exploration, the algorithm samples from the distribution, and then performs a distance or distribution based filter.

Following the approach in Algorithm \ref{alg:dis}, we intuitively extend the approach to the continuous case, where the number the number of components in the gaussian mixture model is akin to the discrete set. In Algorithm \ref{alg:bgm_cf} we also include a filter to guarantee that the obtained sampled is unlikely to be from the original gaussian distribution.

\begin{algorithm}[H]
\label{alg:bgm_cf}
\caption{Continuous Counterfactual via Bayesian Gaussian Mixture}
\begin{algorithmic}[1]
\STATE \textbf{Input:} Feature index $j$, feature vector $\mathbf{x}$, original value $v$, BGM model $\mathcal{G}$ with $k$ components
\STATE \textbf{Output:} feature vector $\mathbf{x'}$ with counterfactual perturbation
\STATE Compute posterior probability given $\gamma_k(v)$
\STATE Identify component: $k^* \leftarrow \arg\max_k \gamma_k(v)$
\STATE Create perturbed model $\mathcal{G}'$ without component $k^*$
\STATE Sample $v' \sim \mathcal{G}'$
\STATE Accept with probability:
\STATE \hspace{1em} $p_{\text{acc}} \leftarrow \min(1, 1 - p_k(v')$
\RETURN $\mathbf{x}$
\end{algorithmic}
\end{algorithm}

\section{Ablation Studies} 

\subsection{Aggregation}

On this ablation study we evaluate the aggregation method over multiple conterfactual samples. For a given sampling budget, then the obtained Causal effect for feature can be aggregate using maximum, mean or median. Figure \ref{fig:aggregation_violin} seems to indicate that taking the maximum value for each feature yields the best results. While the results are all similar, as expected, using the maximum value has better results across both metrics. 

We also can observe the different aggregation methods across different sampling budgets. To perform an extensive and robust analysis, 5 independent evaluation procedures are performed at each sampling budget, from 1 to 100, at spaced intervals. As expected, increasing the number of samples improves the quality of explainer. We can also conclude that at around 10 samples, the improvements are less significant. Finally, it is clear that using the maximum as the aggregation function is superior for all number of samples.

\begin{figure}[htbp]
    \centering
    \begin{subfigure}{0.49\textwidth}
        \centering
       \includegraphics[width=\textwidth]{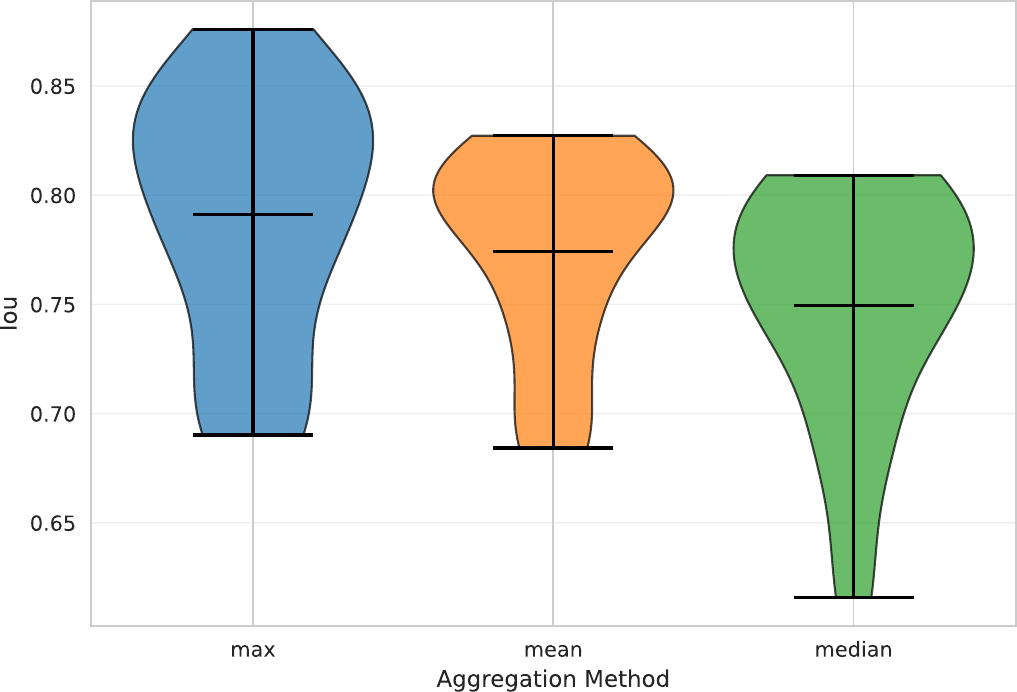}
        \caption{IoU}
    \end{subfigure}
    \begin{subfigure}{0.49\textwidth}
        \centering
        \includegraphics[width=\textwidth]{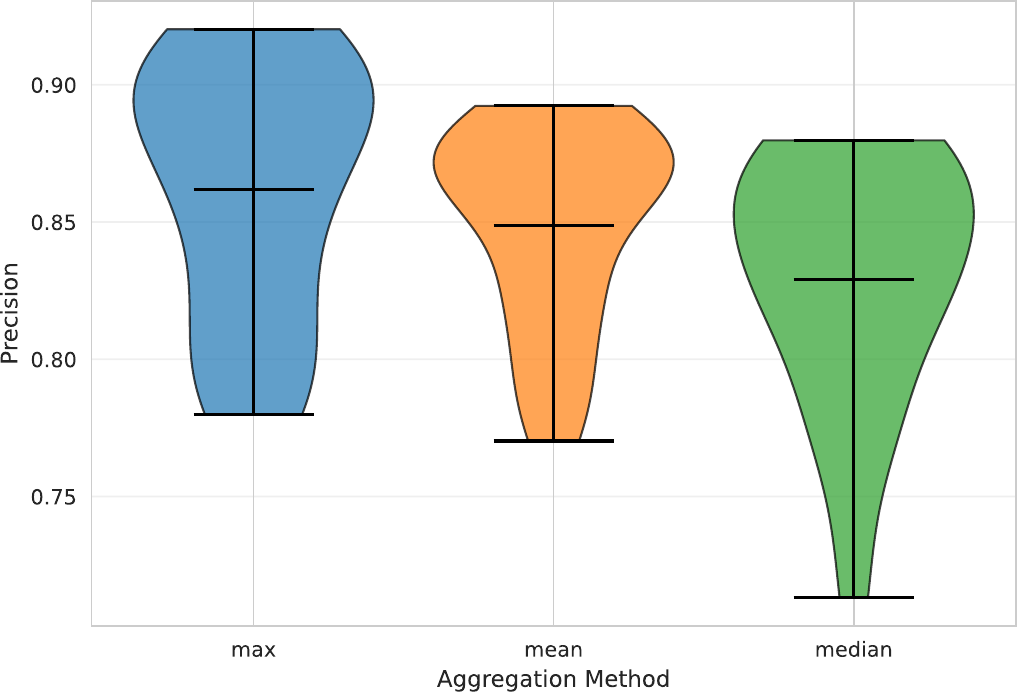}
        \caption{Precision}
    \end{subfigure}
        \caption{Comparison of aggregation method results for Causal Effect. The values are calculated across all classification models, for all datasets in the node classification task.}
    \label{fig:aggregation_violin}
\end{figure}

\begin{figure}[htbp]
    \centering
    \begin{subfigure}{0.49\textwidth}
                \centering
               \includegraphics[width=0.8\textwidth]{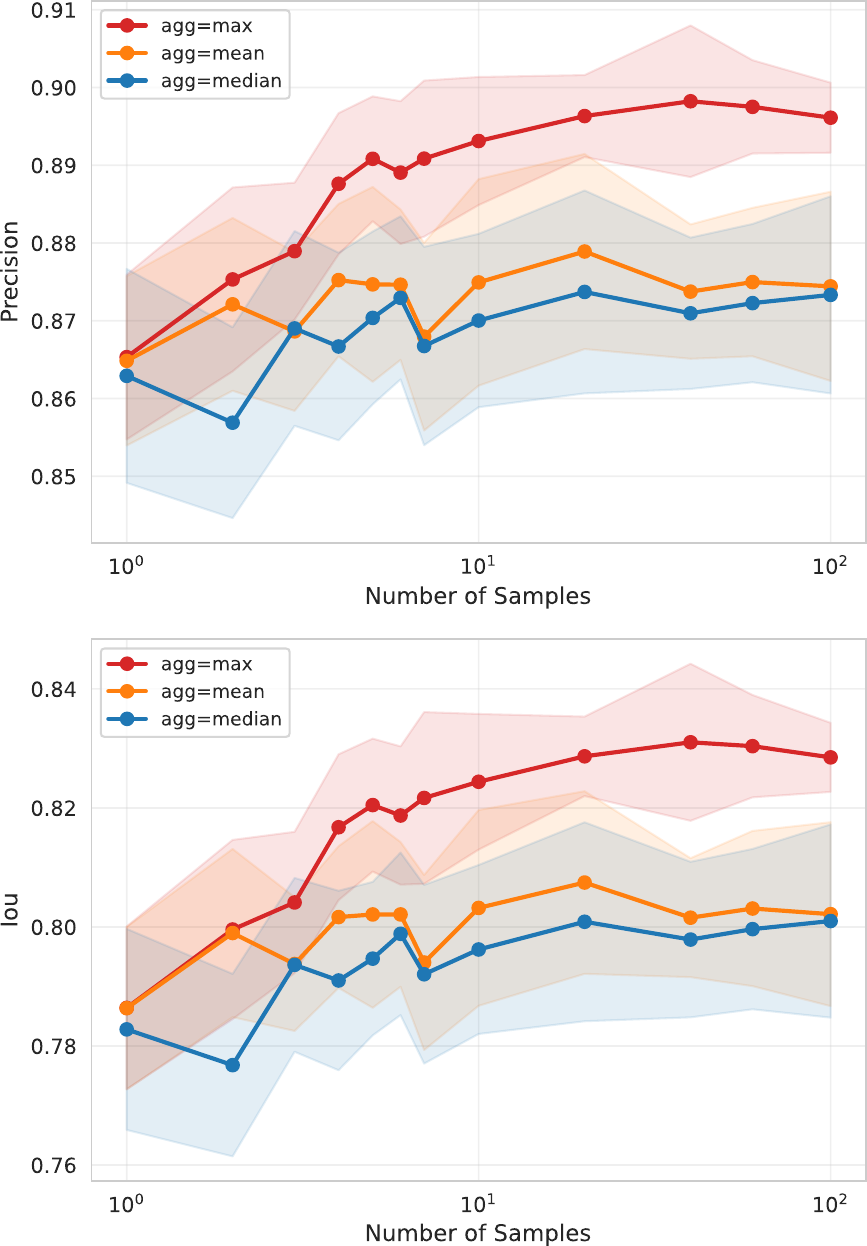}
               \caption{Tree-Grid Dataset}
    \end{subfigure}
    \begin{subfigure}{0.49\textwidth}
        \centering
        \includegraphics[width=0.8\textwidth]{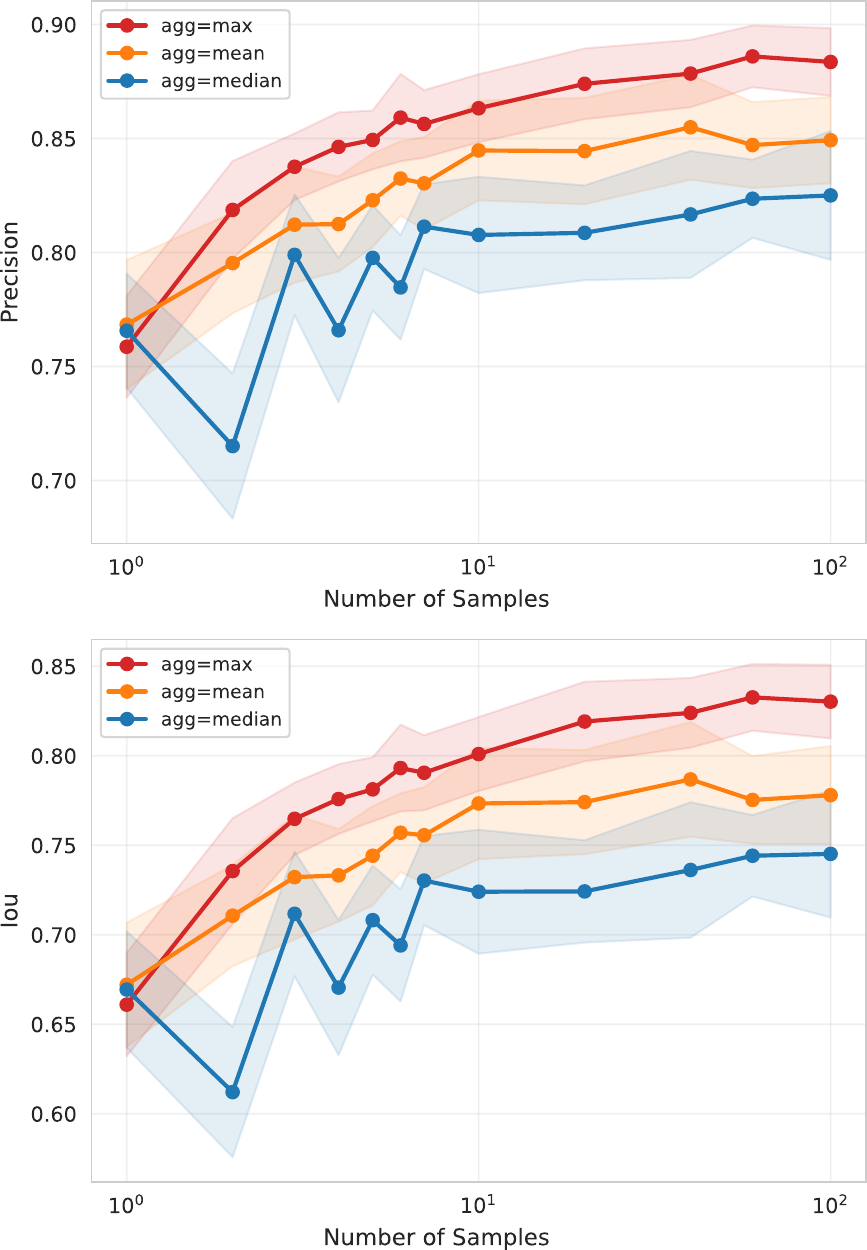}
        \caption{BA-Shapes}
    \end{subfigure}
        \caption{Comparison of aggregation method results for Causal Effect. The values are calculated across all classification models, for all datasets in the node classification task.}
        \label{fig:agg_samples_cf}
\end{figure}

\newpage
\subsection{Conterfactual Distribution}

As mentioned in \ref{sec:cont_feat}, there are different distributions from which to sample counterfactuals. By choosing a specific distribution, it implies assumptions about the underlying data distribution and, consequently, affects the properties of the generated counterfactuals. This section presents an ablation study on the impact of this choice. To do so, we evaluate different counterfactual distributions ($\mathcal{G}$) from which to sample from, for the datasets with continuous features.

\begin{figure}[htbp]
    \centering
    \begin{subfigure}{0.49\textwidth}
                \centering
               \includegraphics[width=0.8\textwidth]{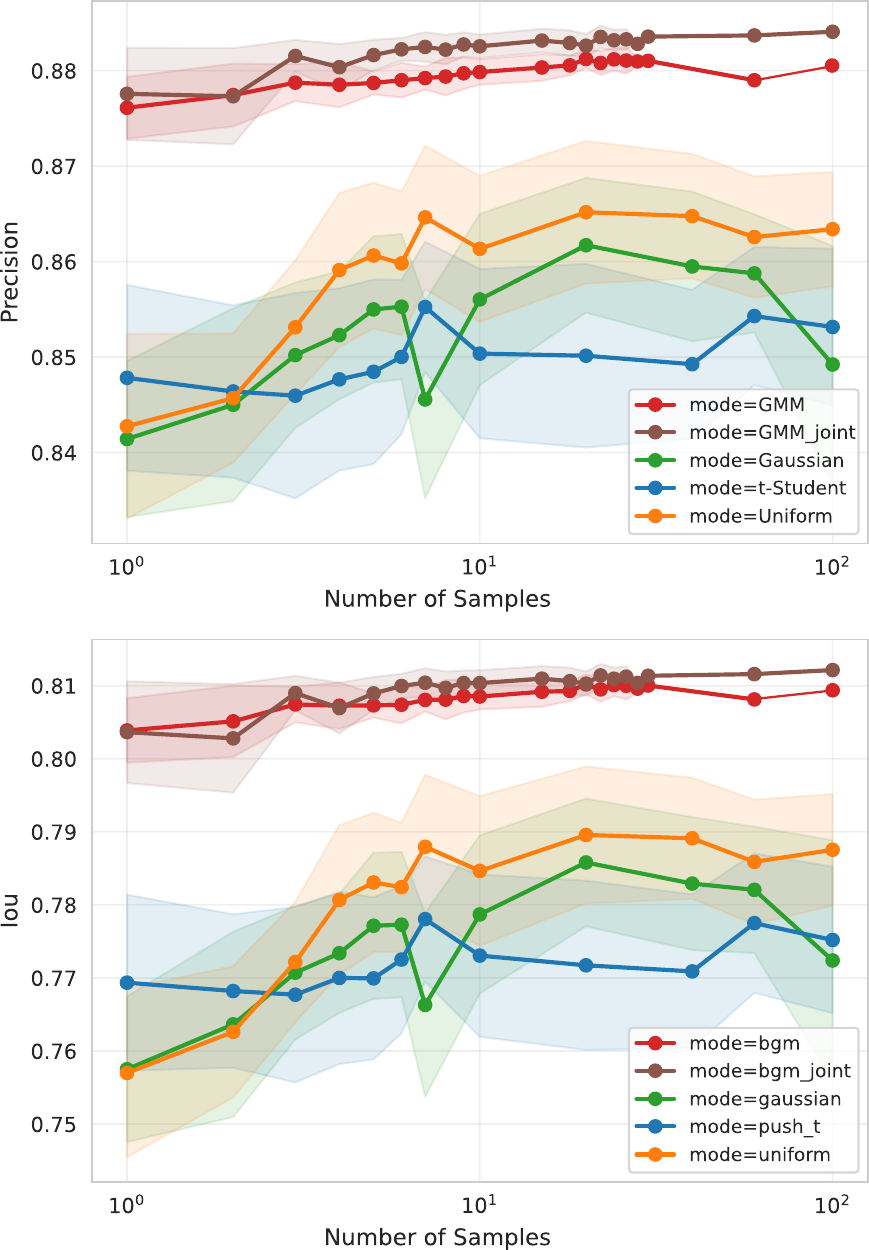}
               \caption{Tree-Grid Dataset}
    \end{subfigure}
    \begin{subfigure}{0.49\textwidth}
        \centering
        \includegraphics[width=0.8\textwidth]{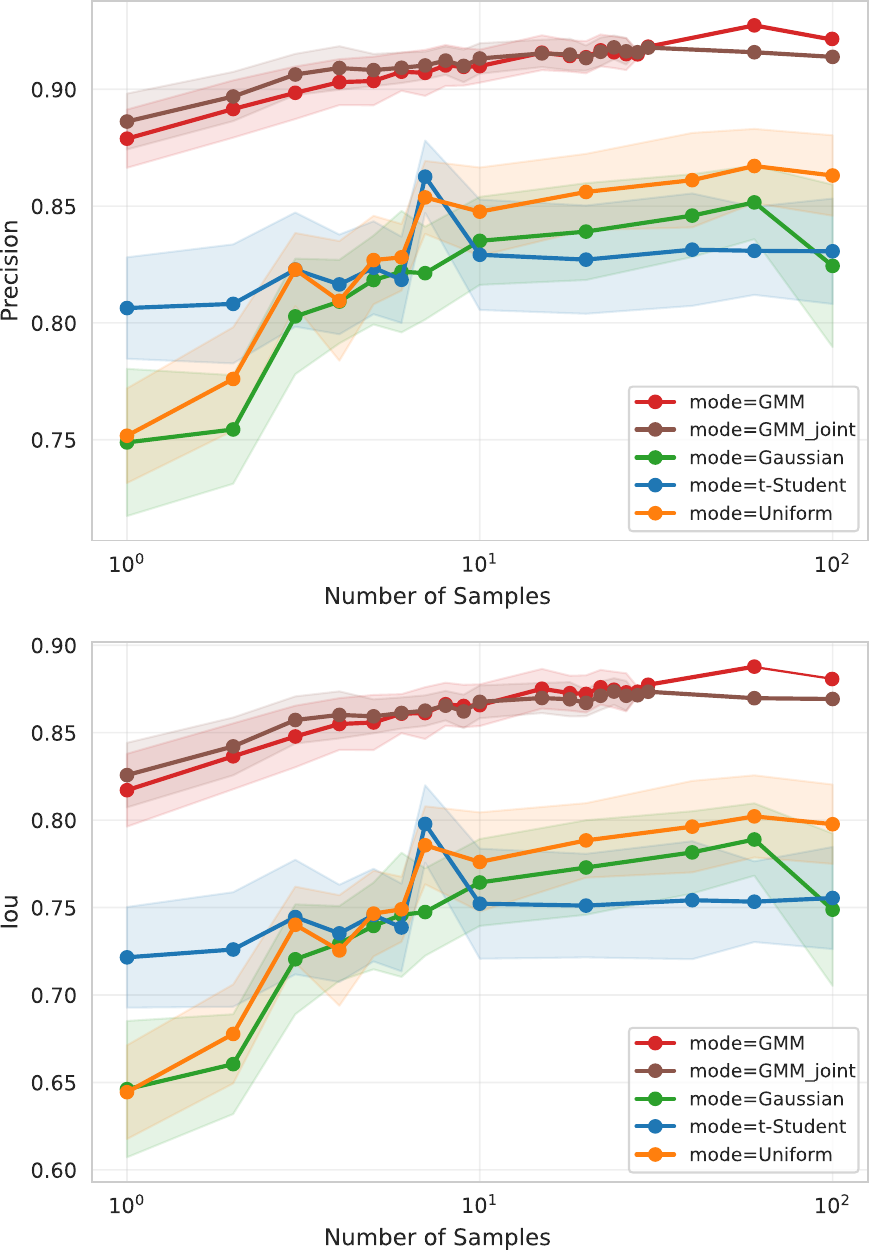}
        \caption{BA-Shapes}
    \end{subfigure}
        \caption{Comparison of sampling distributions results for Causal Effect. The values are calculated across all classification models, for all datasets in the node classification task.}
        \label{fig:agg_samples_cf_mode}
\end{figure}

In Figure \ref{fig:agg_samples_cf} it is clear that the Gaussian Mixture Model is a better counterfactual model than the simpler gaussian, uniform or t-student, for a fixed number of samples, using the maximum as the aggregation function. Figure \ref{fig:agg_samples_cf_mode} further reinforces this observation, by indicating that the GMM is better across all samplings budgets, including $N=1$. This shows that the GMM is able, with some success, to correctly represent the feature space. In this ablation study we also evaluate two different approaches, sampling from the marginal distribution and sampling from the joint distribution. For the given datasets, the results are very similar, with some slight advantage to the joint approach, as expected. While we do not extrapolate this finding to more complex feature spaces, it does shows that, for this cases, sampling from the marginal is an acceptable simplification that can be applied with very limited downside.

\begin{figure}[htbp]
    \centering
    \begin{subfigure}{0.49\textwidth}
        \centering
       \includegraphics[width=\textwidth]{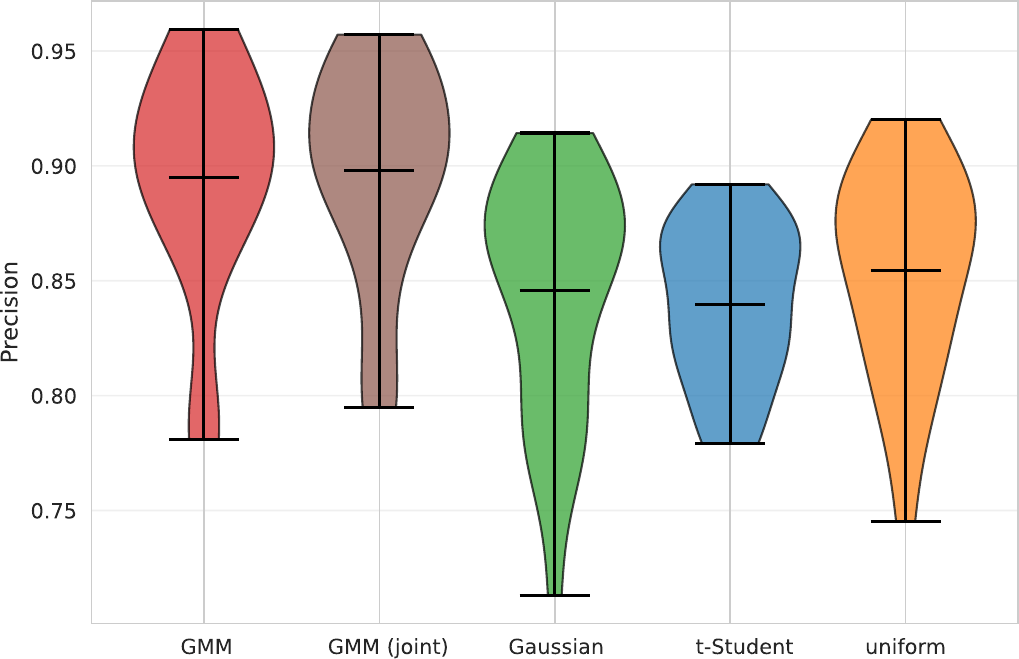}
        \caption{IoU}
    \end{subfigure}
    \begin{subfigure}{0.49\textwidth}
        \centering
        \includegraphics[width=\textwidth]{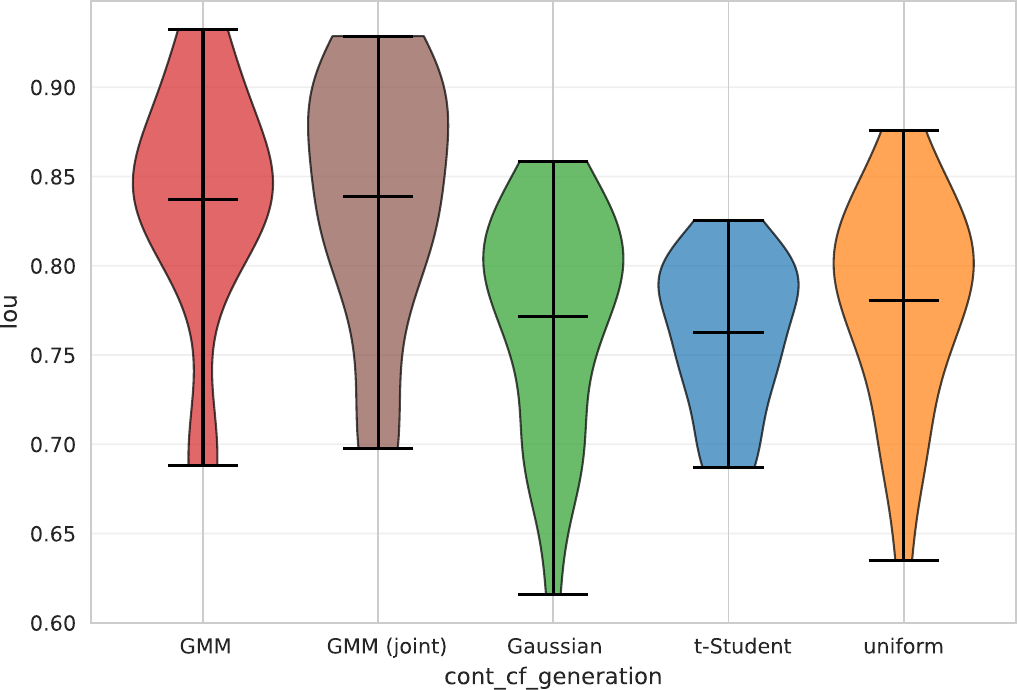}
        \caption{Precision}
    \end{subfigure}
        \caption{Comparison of aggregation method results for Causal Effect. The values are calculated across all classification models, for all datasets in the node classification task.}
    \label{fig:aggregation_violin}
\end{figure}

\newpage
\section{Graph Neural Networks Training Results}
\label{sec:training_appendix}

The main objective of this paper is to propose a causal explanation method for GNNs, that achieves both high explainability but is also interpretable, in the sense that someone can understand the finding of the explanation model. To do this, we trained four different GNNs models for each dataset, using very simple datasets with ground truth labels. While there are more realistic and complex datasets, those datasets lack proper ground-truth label, and the evaluation of the explainers becomes intertwined with the predictive capabilities of the model themselves.

The datasets chosen are commonly used for GNN explainability, and due to their simplicity allow to isolate explainability component. For fairness, we fix the overall model architecture and all hyperparameters within each task setting, with the only difference being the type of layer used in the model. For each dataset, we trained a GCN, a GraphSAGE, a GAT, and a GIN model, using pytorch geometric \cite{Fey/etal/2025}. Table \ref{tab:gnn_models} shows the configuration of GNNs models for each task. For all datasets, we adopt a random 80/10/10 split for training, validation, and testing, respectively.


\begin{table}[htbp]
  \centering
  \caption{GNN model configuration per task.}
  \label{tab:gnn_models}
  \vspace{2mm}
    \begin{tabular}{cccc}
        \toprule
        & Node Classification & Graph Classification  \\
      \midrule
      \textbf{\# Layers} & 3 & 3  \\ 
      \midrule
      \textbf{Hidden Dimension} & 20 & 20  \\
      \midrule
      \textbf{Pooling Layer} & - & Max  \\
      \midrule
      \textbf{Criterion} & CE or BCE & BCE  \\
      \midrule
      \textbf{Optimizer} & Adam & Adam  \\ 
      \midrule
      \textbf{Learning Rate} & $1e^{-3}$ & $1e^{-3}$  \\ 
      \midrule
      \textbf{\# Epochs} & 1000 & 1000  \\ 
      \midrule
      \textbf{Train/Val/Test Split} & 80/10/10 & 80/10/10  \\ 
      \bottomrule
    \end{tabular}%
    
\end{table}

\subsection{Node Classification}

\begin{figure}[htbp]
    \centering
    \caption{Training Loss for the node classification task. Loss is presented for the training and validation sets. Across the models we consistently observe GraphSage as the fastest and more consistent model to achieve close to zero loss, while GIN fails to properly optimize in these tasks.}
    \label{fig:nc_loss}
    
    \begin{subfigure}{0.49\textwidth}
        \centering
       \includegraphics[width=\textwidth]{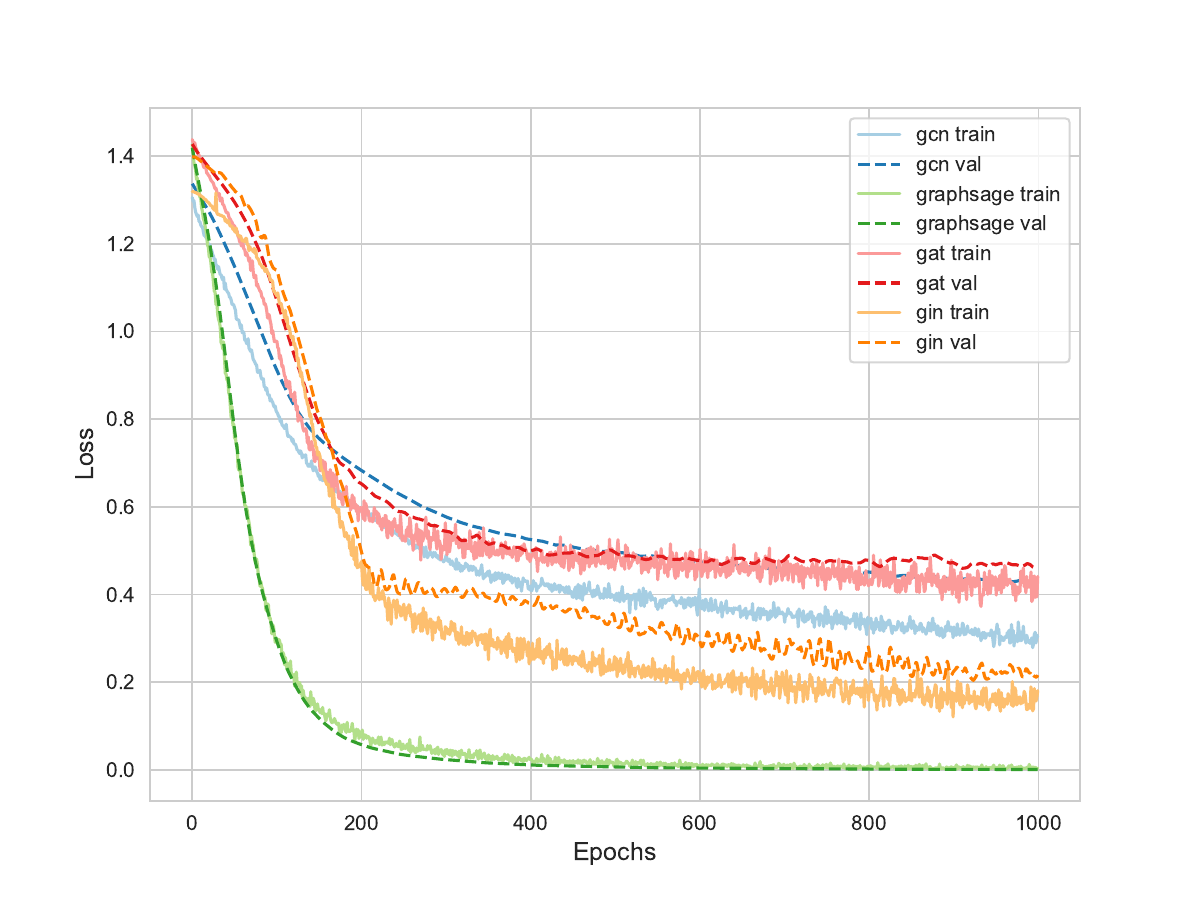}
        \caption{BA-Shapes}
    \end{subfigure}
    \begin{subfigure}{0.49\textwidth}
        \centering
        \includegraphics[width=\textwidth]{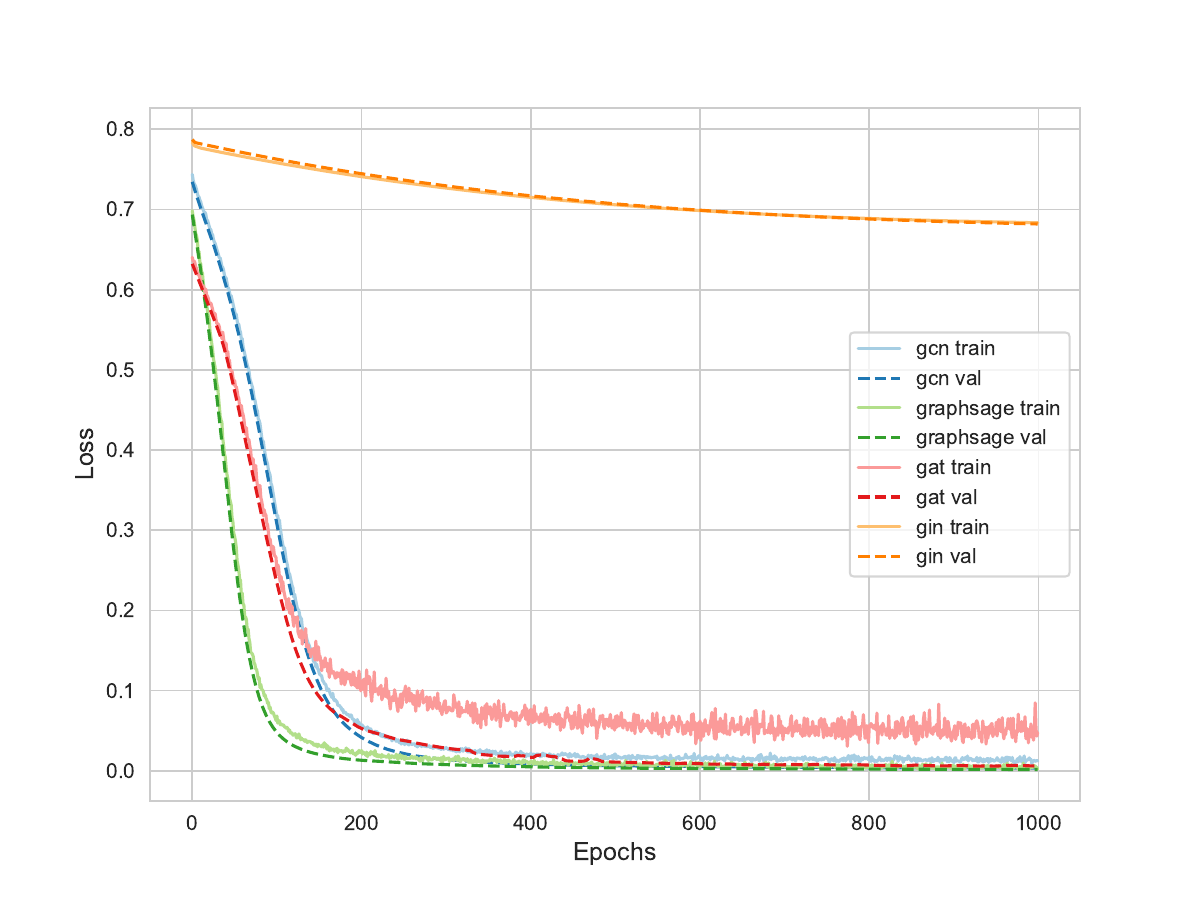}
        \caption{Tree-Grid}
    \end{subfigure}
\end{figure}

 We measure model training performance on all datasets using the loss. As shown in Fig.~\ref{fig:nc_loss}, GraphSAGE exhibits the fastest convergence and consistently achieves the lowest final training loss. In contrast, GIN fails to converge on the Tree-Grid dataset. This finding is complemented in Table \ref{tab:nc_test_metrics}, where the graphSAGE models achieves perfect accuracy in both test sets, and GIN has a subpar accuracy of $0.5887$. 

\begin{table}[!h]
\centering
\caption{Node classification test metrics.}
\label{tab:nc_test_metrics}
\small
\setlength{\tabcolsep}{4pt}
\begin{tabular}{c|ccc|ccc}
\toprule
& \multicolumn{3}{c|}{\textbf{BA-Shapes}} 
& \multicolumn{3}{c}{\textbf{Tree-Grid}} \\
\cmidrule(lr){2-4} \cmidrule(lr){5-7}
\textbf{Model} 
& Loss & Acc & F1 
& Loss & Acc & F1 \\
\midrule
GCN 
& 0.3606 & 0.9000 & 0.8689 
& 0.0083 & \textbf{1.0000} & \textbf{1.0000} \\

GraphSAGE 
& \textbf{0.0005} & \textbf{1.0000} & \textbf{1.0000} 
& \textbf{0.0033} & \textbf{1.0000} & \textbf{1.0000} \\

GAT 
& 0.4138 & 0.8286 & 0.7423 
& 0.0195 & 0.9839 & 0.9832 \\

GIN 
& 0.1733 & 0.9571 & 0.9427 
& 0.6827 & 0.5887 & 0.3706 \\
\bottomrule
\end{tabular}
\end{table}

\subsection{Graph Classification}

\begin{figure}[htbp]
    \centering
    \caption{Training accuracy for the graph classification task. Accuracy is presented for the training and validation sets.}
    \label{fig:gc_loss}
    
    \begin{subfigure}{0.49\textwidth}
        \centering
       \includegraphics[width=\textwidth]{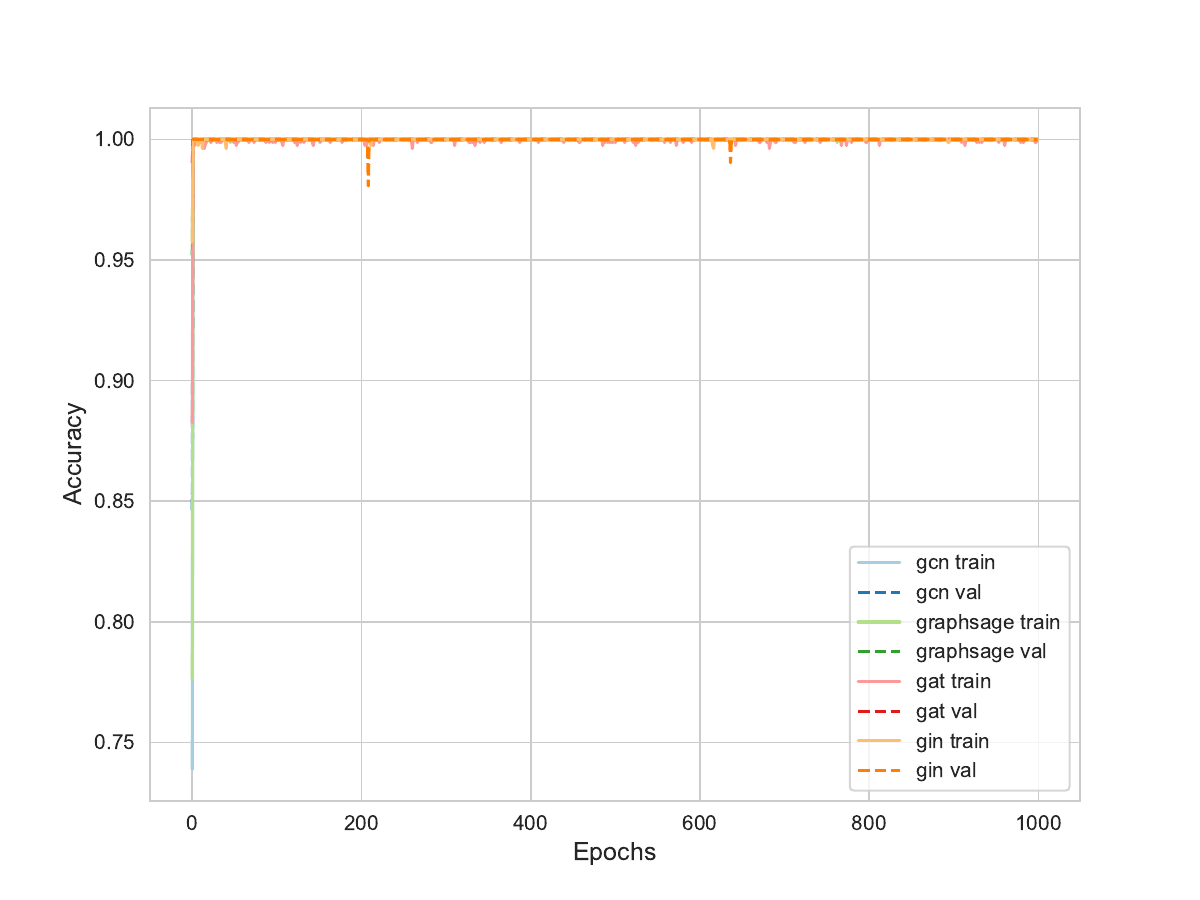}
        \caption{BA-2Motif}
    \end{subfigure}
    \begin{subfigure}{0.49\textwidth}
        \centering
        \includegraphics[width=\textwidth]{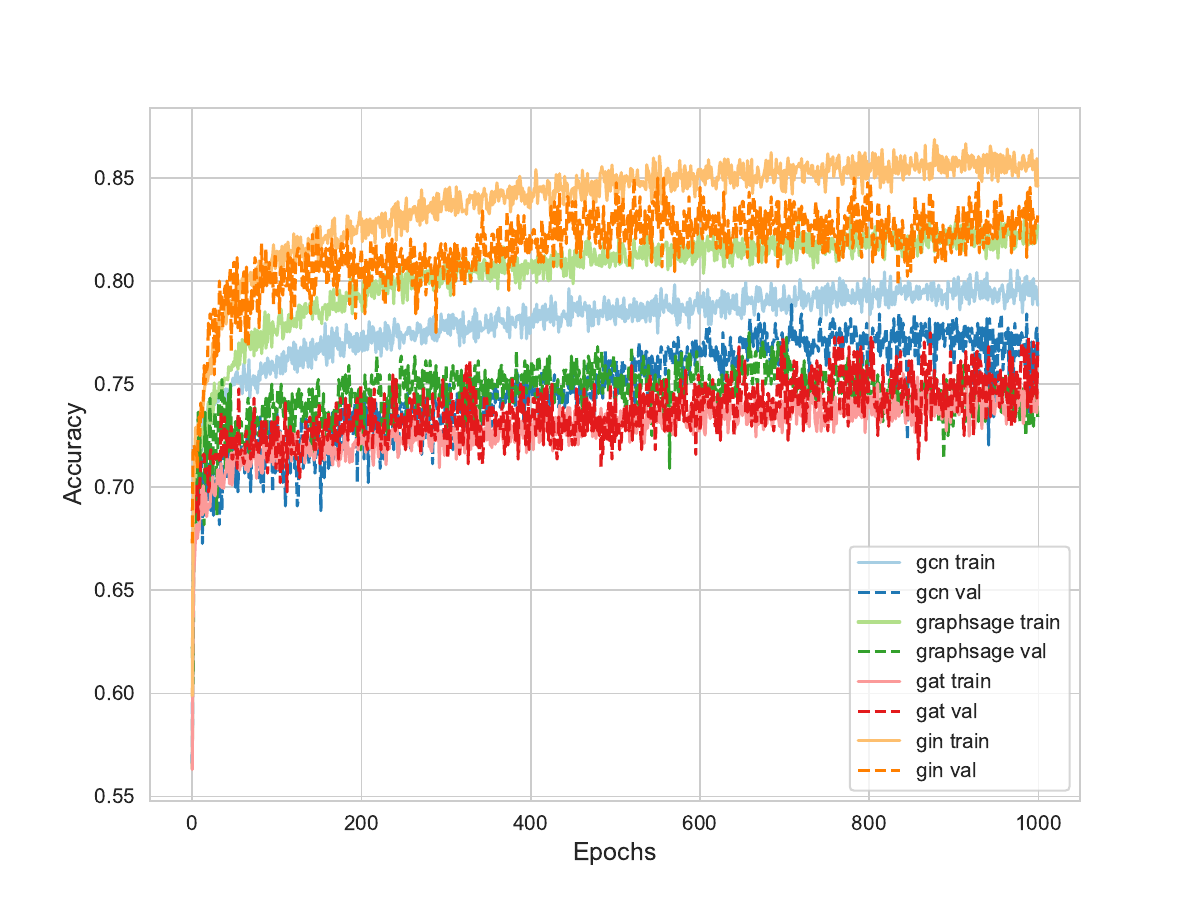}
        \caption{MUTAG}
    \end{subfigure}
\end{figure}

The results for graph classification are similar to node classification in the synthetic graph classification dataset BA-2motifs in Fig. \ref{fig:gc_loss} and Table \ref{tab:gc_test_metrics}. The perfect accuracy results are to be expected, since the features of these datasets are specially created to make it easier for the model to learn and predict. GraphSAGE again has the lowest loss on the test set, and trains the fastest.

\begin{table}[htbp]
\centering
\caption{Graph classification test metrics.}
\label{tab:gc_test_metrics}
\small
\setlength{\tabcolsep}{4pt}
\begin{tabular}{c|ccc|ccc}
\toprule
& \multicolumn{3}{c|}{\textbf{BA-Shapes}} 
& \multicolumn{3}{c}{\textbf{Tree-Grid}} \\
\cmidrule(lr){2-4} \cmidrule(lr){5-7}
\textbf{Model} 
& Loss & Acc & F1 
& Loss & Acc & F1 \\
\midrule
GCN 
& 0.0000 & 1.0000 & 1.0000
& 0.4972 & 0.7894 & 0.7461 \\

GraphSAGE 
& 0.0000 & 1.0000 & 1.0000 
& 0.6229 & 0.7545 & 0.6865 \\

GAT 
& 0.0002 & 1.0000 & 1.0000 
&  0.5705 & 0.7227 & 0.6841 \\

GIN 
& 0.0000 & 1.0000 & 1.0000 
& \textbf{0.4321} & \textbf{0.8182} & \textbf{0.7958} \\
\bottomrule
\end{tabular}
\end{table}

For the MUTAG dataset, we can see in Fig. \ref{fig:gc_loss} that the models learn more slowly and do not perform as well on the validation set. The reason is that the features in the MUTAG dataset are not specifically designed to help the model learn and make predictions more easily. Therefore, the quality of the predictions is more dependend on the GNN aggregation layer. Even so, for the MUTAG dataset, GIN is the GNN that performs best on the training and validation sets,  with the least overfitting. The test performance in Table \ref{tab:gc_test_metrics} shows that the models generalize well to unseen data but do not achieve perfect metric scores. This is expected since the MUTAG dataset is more complex and does not have a feature set designed to facilitate model learning and prediction, and it can influence the explanation evaluation, since it is not expected that a model that wrongly classifies the graph will have a correct explanation subgraph.

\section{G2TeXplainer Extra Evaluation}
\label{app:g2t_extra_eval}

\begin{table*}[th]
\centering
\caption{Evaluation of G2TeXplainer against GraphXAIN on the full test set (5 samples per example). Judge-based metrics are scored from 1 (very poor) to 5 (perfect). \textbf{N. Fid.}: correctness of node importance; \textbf{Str.}: accuracy of detected motifs and node memberships; \textbf{Clar.}: coherence and interpretability of the explanation; }
\label{tab:g2texplainer_llm_judge}
\footnotesize
\setlength{\tabcolsep}{2.7pt}
\renewcommand{\arraystretch}{1.05}
\begin{tabular}{lccc}
\toprule
 \textbf{Prompt} 
& \textbf{N. Fid.} 
& \textbf{Str.} 
& \textbf{Clar.} 
\\
\midrule
GraphXAIN & 3.568 & 3.296 & 3.762 \\
P4  Motif-Aware + R. & 4.100 & 3.882 & 3.436 \\
\end{tabular}
\end{table*}

\begin{figure}[htbp]
    \centering
    \caption{Distribution of LLM-as-judge Node Fidelity scores for different graph types. Each cell shows the number of explanations (and percentage) receiving a specific hallucination score from 1 (low Fidelity) to 5 (high Fidelity).}
    \includegraphics[width=0.9\textwidth]{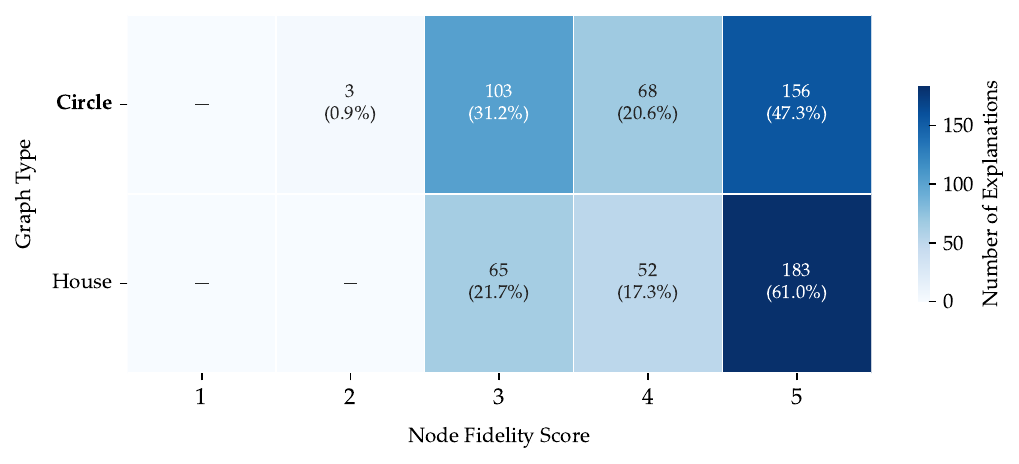}
    \label{fig:gcn_hallu_score_distribution}
\end{figure}

\begin{figure}[htbp]
    \centering
    \caption{Distribution of LLM-as-judge Structural scores for different graph types. Each cell shows the number of explanations (and percentage) receiving a specific structural score from 1 (poor structure) to 5 (well-structured).}
    \includegraphics[width=0.9\textwidth]{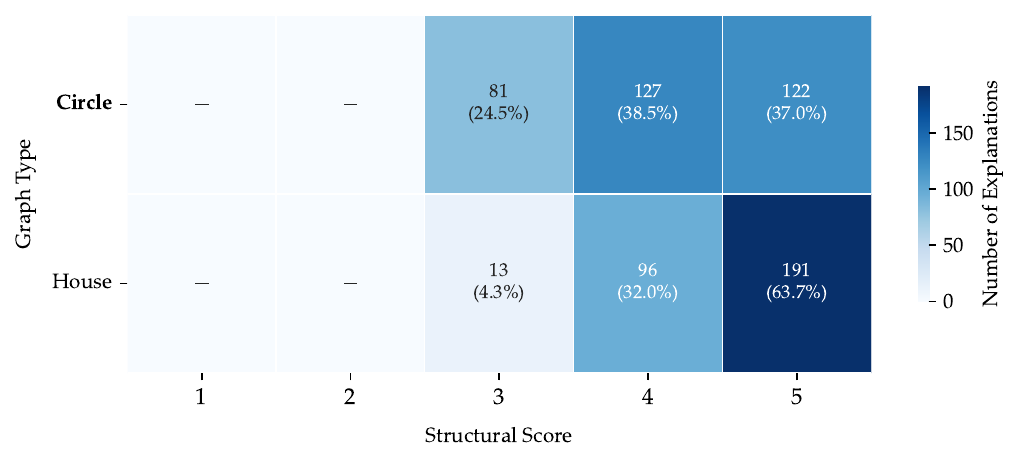}
    \label{fig:gcn_struct_score_distribution}
\end{figure}

\section{G2TeXplainer P4 Prompt}

\begin{tcolorbox}[colback=pink!5, colframe=black, title=Fine-tuned prompt used for explanation description.]
\begin{verbatim}
You are analyzing why a Graph Neural Network (GNN) identified specific nodes
as important for its prediction.

TASK:

You are NOT asked to explain why the graph is a "{label}".
You are asked to explain whether the EXPLAINABLE NODES support that 
prediction.

Rules:
Only discuss nodes in {explain_nodes}.
Do NOT discuss nodes that were not selected.
Do NOT explain the full graph structure.
Focus only on whether the selected nodes align with the "{label}" motif.

Instructions:

Identify which of the selected nodes belong to the listed structures.
If selected nodes are part of a listed 4-cycle or 5-cycle, explain how that 
supports the "{label}" pattern.
If selected nodes are isolated or not part of the listed structures, state 
that they do NOT support the prediction.
Assess whether the explanation nodes meaningfully justify the "{label}"
prediction.
STRUCTURAL DEFINITIONS:
CIRCLE PATTERN:
Definition: A connected 5-node cycle
HOUSE PATTERN:
Definition: 4-cycle with triangle
NONE PATTERN:
The graph does not contain either the Circle or House structural motifs.
GNN PREDICTION: "{label}"
EXPLAINABLE NODES: {explain_nodes}

STRUCTURES DETECTED IN SUBGRAPH:
{structure_text}

Write 4–6 analytical sentences.
Explanation:

\end{verbatim}
\end{tcolorbox}
\end{document}